%% file: paper.tex
\pdfoutput=1
\documentclass[preprint]{elsarticle}
\usepackage[latin1]{inputenc}
\usepackage{amssymb,amsmath,array}
\usepackage{bm}
\newtheorem{theorem}{Theorem}
\usepackage{float}
\usepackage{microtype}
\usepackage{todonotes}
\usepackage{multirow}
\usepackage{graphicx}
\usepackage{subcaption}
\usepackage{hyperref}
\usepackage{epstopdf}
\input{mymath}

\newdefinition{rmk}{Remark}
\newdefinition{definition}{Definition}

\makeatletter
\def\ps@pprintTitle{%
    \let\@oddhead\@empty
    \let\@evenhead\@empty
    \def\@oddfoot{\footnotesize\itshape
      {Preprint accepted at Neurocomputing, \href{https://creativecommons.org/licenses/by-nc-nd/4.0/}{CC-BY-NC-ND 4.0 license}} \hfill\today}%
    \let\@evenfoot\@oddfoot
    }
\makeatother

\begin{document}
\title{Feature Relevance Determination for Ordinal Regression in the Context of Feature Redundancies and Privileged Information\tnoteref{t1}}
\tnotetext[t1]{Funding by the DFG in the frame of the graduate school DiDy (1906/3) and by the BMBF (grant number 01S18041A) is gratefully acknowledged.}
\author[1]{Lukas Pfannschmidt\corref{cor1}}
\ead{lukas@lpfann.me}
\cortext[cor1]{Corresponding author}
\author[1]{Jonathan Jakob}
\ead{jjakob@techfak.uni-bielefeld.de}
\author[1]{Fabian Hinder}
\ead{fhinder@techfak.uni-bielefeld.de}
\author[2]{Michael Biehl}
\ead{m.biehl@rug.nl}
\author[3]{Peter Tino}
\ead{p.tino@cs.bham.ac.uk}
\author[1]{Barbara Hammer}
\ead{bhammer@techfak.uni-bielefeld.de}

\address[1]{Machine Learning Group, Bielefeld University, DE}
\address[2]{Intelligent Systems Group, University of Groningen, NL}
\address[3]{Computer Science, University of Birmingham, UK}

\begin{abstract}
Advances in machine learning technologies have led to increasingly powerful models in particular in the context of big data.
Yet, many application scenarios demand for robustly interpretable models rather than optimum model accuracy; as an example, this is the case if potential biomarkers or causal factors should be discovered based on a set of given measurements.
In this contribution, we focus on feature selection paradigms, which enable us to uncover relevant factors of a given regularity based on a sparse model.
We focus on the important specific setting of linear ordinal regression, i.e.\ data have to be ranked into one of a finite number of ordered categories by a linear projection.
Unlike previous work, we consider the case that features are potentially redundant, such that no unique minimum set of relevant features exists.
We aim for an identification of all strongly and all weakly relevant features as well as their type of relevance (strong or weak); we achieve this goal by determining feature relevance bounds, which correspond to the minimum and maximum feature relevance, respectively, if searched over all equivalent models.
In addition, we discuss how this setting enables us to substitute some of the features, e.g.\ due to their semantics, and how to extend the framework of feature relevance intervals  to the setting of privileged information, i.e.\ potentially relevant information is available for training purposes only, but cannot be used for the  prediction itself.
\end{abstract}


\maketitle

\section{Introduction}
Ordinal regression refers to the task to assign data to a finite number of classes or bins, which are ordered qualitatively along a preference scale.
Ordinal data often occur in sociodemographic, financial or medical contexts where it is difficult to give absolute quantitative measurements but easily possible to compare samples and assign those to different bins, which are  qualitatively ordered, such as the severity of a disease or the risk of a financial transaction.
Another popular example for a ranking on ordinal scales takes place in customer feedback or product ranking by humans~\cite{harper2016movielens}.
Here, the quality is often represented by a five-star rating scale, where five stars correspond to the best rating and one star to the worst.
Indeed, many  human ratings  are represented in an ordinal scale rather than absolute values.

The {\it ordinal regression problem}~(ORP) is the task to embed given data in the real numbers such that they are ordered according to their label, i.e.\ the target bin. 
An error is encountered whenever an ordering of two data points assigned to different bins is violated.
Although the problem can be treated as a regular regression or classification method,
dedicated techniques are often preferred, since they can account for the fact that the
distance between ordinal classes in the data is unknown and not necessarily evenly distributed.
Examples of ordinal regression include
treatments such as the multiclass classification problem~\cite{FrankSimpleApproachOrdinal2001},
and extensions of standard models such as the support vector machine (SVM) or learning vector quantization (LVQ) to ordinal regression tasks
\cite{ShashuaRankingLargeMargin,svmo,pOGMLVQ,tino}.
A recent work proposed an incremental and sparse Bayesian approach with favourable scaling properties~\cite{li2018a}.
Often, ordinal regression is treated as a pairwise ranking problem
\cite{DBLP:journals/corr/abs-1901-07884}.
Further, there does exist recent theoretical work which establishes consistency of some surrogate losses for ordinal regression, which have better numeric properties~\cite{DBLP:journals/corr/PedregosaBG14}.

In this work, we will rely on SVM-like treatments of the ORP due to the mathematical elegance and flexibility of this formulation~\cite{ShashuaRankingLargeMargin,svmo,pOGMLVQ}.

Recently, methods which enable the interpretability of machine learning models have extensively been discussed~\cite{Guidotti:2018:SME:3271482.3236009}. One common way to enhance model interpretability is by means of a determination of the most relevant input dimensions or features, i.e.\ the relevance of ordinal explanatory variables for the given task. This is particularly relevant when the objective exceeds mere diagnostics, such
as safety-critical decision-making, or the design of repair strategies.
There do exist a few approaches which address such feature selection for  ordinal regression:
The approach~\cite{geng2007} uses a minimal redundancy formulation based on a feature importance score to find the subset of relevant features.
The work in~\cite{baccianella2010} focuses on multiple filter methods which are  adapted to ranking data.
These models deliver sparse ordinal regression models which enable some insight into the underlying classification prescription. Yet, their result is arbitrary in the case of correlated or redundant features: if there does not exist a unique minimum relevant feature set, it often depends on arbitrary initialization or algorithmic design choices,  which feature from a set of redundant features is chosen.  Hence, possibly relevant features, so-called weakly relevant features, can easily be overlooked, albeit they might have a substantial contribution or even causal influence to a model.

The so-called  \emph{all} relevant feature selection problem deals with the challenge to determine all features, which are potentially relevant for a given task -- a problem which is particularly important for diagnostics purposes if it is not priorly clear which one of a set of relevant, but redundant features to choose.
Finding this subset is generally computationally intractable.
For standard classification and regression schemes, a few  efficient
heuristics have been proposed:
one possibility is to quantify not only the relevance but also the redundancy of features~\cite{Yu:2004:EFS:1005332.1044700}.
Another popular model extends predictive models with statistical tests to discriminate between relevance and irrelevance~\cite{kursa2010}.
Recently, the problem of feature relevance has been investigated in the special case of linear mappings; here, the problem can be phrased in terms of relevance intervals, leading to a convex problem and superior performance in benchmarks~\cite{christina}.
In the presented work, the goal is to extend this
approach to the specific relevant setting of ordinal regression  tasks, and to demonstrate the benefit of this model in comparison to alternative popular feature selection models such as lasso or ElasticNet.

Besides a formal mathematical modelling by means of linear optimization tasks, we will also demonstrate the suitability of
the model to investigate the role of critical features for an ORP.
As an example, the integration of criteria such as age, gender, or ethnicity might improve the prediction accuracy of a given model as measured by an appropriate cost function~-~yet, it might be debatable if these features can have any relevance for the given task as regards a causal relationship on the one hand; on the other hand, it might be  unethical or impossible to actually gather such features for a prediction model in its daily use. Examples for a questionable impact of such characteristics on a formal model have recently been debated under the umbrella of model fairness~\cite{Kearns:2017:FAM:3033274.3084096}.
We will discuss how feature relevance profiles, in particular the identification of weakly relevant features, enable further insight into such settings, by explicitly quantifying the possible impact of such features.

There exists another popular setting where not all features can or should be used for daily use, hence feature relevances are of particular importance: the scenario of so-called privileged information phrases the situation that some features are available during the training phase only, but not during the test phase, e.g.\ due to the costs, computational load, or any other restrictions.
In classical machine learning, it is commonly assumed, that training and test set have an identical statistical distribution and utilize the same predictive features.
In contrast, the \emph{learning using privileged information} paradigm (LUPI)~\cite{vapnik2009} considers additional privileged information only available at training time.
This paradigm can be understood as an intelligent teacher feeding the learner extra information to improve the learning process~\cite{vapnik2015}.
Additional information could be the output of another model (`machines-teaching-machines') or input from a human expert itself, who intuitively knows which examples in the data are hard to discriminate.
Examples are medical measurements which require invasive techniques or measurements which require too much time in daily use, but would be affordable for training.
The approach~\cite{vapnik2009} proposed a variant of SVMs that incorporates privileged information for training.
The modelling replaces or enriches slack variables, which are required by soft-margin SVMs to correct for hard training samples.
This specific approach is known as \emph{similarity control}~\cite{vapnik2015}.
The approach~\cite{vapnik2009} introduces the SVM+ in which  a smooth function based on the privileged information (PI) is used at training time to improve learning in non-separable classification settings.
The method~\cite{tang} refrained from fully replacing the slack variables and combined them with a smooth function based on PI. It achieved better generalization ability and lower complexity models.
Furthermore, this approach also extends the SVM+ to ordinal regression problems.

While approaches to incorporate privileged information exist, and it has been shown that LUPI has the potential to speed up learning~\cite{DBLP:conf/nips/PechyonyV10},  the analysis of feature relevances in the context of redundant feature information is still widely open in this setting. In this article, we  also
introduce  an extension of the feature-relevance-interval-computation scheme as proposed in~\cite{christina} to  the LUPI setting;
this addresses the question of which features are potentially relevant to facilitate training, i.e.\ they carry important information to improve the learnability of a task. Irrelevant features in the LUPI framework, on the other hand, do not contribute to the learnability. Unlike standard feature relevances for regression or classification, features relevances for privileged information answer the question whether feature information is beneficial for the learning process itself.

In the following, we will introduce and extend feature relevance learning in the context of redundant features for ordinal regression and privileged information.
For this purpose, we recapture two large margin ordinal regression
formalizations in section~\ref{sec:methods}, which differ in the type of constraints they enforce on ordinal classes, namely \emph{implicit} and \emph{explicit} constraints.
We extend them to an optimization scheme to determine feature relevance bounds in section~\ref{sec:relev_bounds},  which can be transferred to several linear optimization problems (Section~\ref{LP_relev_bounds}).
Further we also define the \emph{explicit} formulation to be used in context of learning using privileged information in Section~\ref{sec:lupi}.
In Section~\ref{sec:experiments} we do several benchmarks to highlight the accuracy and feature selection performance in the classical machine learning case.
In Section~\ref{sub:experiment_privileged_information} we repeat this in the LUPI setting where we focus on performance measures split by the regular and privileged feature set.

\section{Large Margin Ordinal Regression}\label{sec:methods}
We consider the following  ordinal regression learning task:
We assume class labels $L=\{1,2,\ldots,l\}$, which are ordered; w.l.o.g.\ we represent those as natural numbers. We assume training data are given, $X=\{\mathbf x_i^j\in\mathbb{R}^n\:|\: i=1,\ldots,m_j,\, j\in L\}$ where data point $x_i^j$ is assigned the class label $j\in L$, i.e.\ $x_i^j$ is contained in bin number $j$.
The full data set has size $m:=m_1+\ldots+m_l$.
Here the index $j$ refers to the ordinal target variable  the  data point $x_i^j$  belongs to.
The ORP can be phrased as the search for a mapping  $f:\mathbb{R}^n\to\mathbb{R}$, which preserves the ordering of bins as indicated by the label information. That means the inequality
$f(\mathbf x_{i_1}^{j_1})<f(\mathbf x_{i_2}^{j_2})$ should hold for all pairs of class labels $j_1<j_2$ and data indices $i_1$ and $i_2$ in these bins.

In the following,
we will restrict to the case of a linear function, i.e.\ $f(\mathbf x) = \mathbf w^{\top}\mathbf x$ with parameter $\mathbf w\in\mathbb{R}^n$.
In particular in the case of high dimensional data such a linear prescription is often sufficient to model the underlying regularity. Further, it enables a particularly strong link of feature relevances and underlying model, as already elaborated in popular sparse models such as lasso~\cite{tibshirani_regression_1996}.
There do exist different possibilities to model the ORP learning problem. 
Here, we will introduce two existing optimization problems, which rely on large margins, and which treat the inequality constraints in two different ways.

\paragraph*{Explicit Order Constraints}
One way to model ordinal regression is by an embedding of data in the real numbers via $f$, whereby the bins are separated by adaptive thresholds $b_j$, which are learned accordingly.
A popular formulation which is inspired by support vector machines imposes a margin around all thresholds $b_j$ for this embedding~\cite{svmo}:

\begin{equation}
  \label{eq:svm_ordreg_exc}
  \begin{aligned}
    \min_{\mathbf w, \mathbf b,\bm \chi,\bm \xi} && \frac{1}{2} \|\mathbf w\|_1 + C \sum_{i,j}\left(\chi_i^j+\xi_i^j\right) \\
  \end{aligned}
\end{equation}
\begin{equation}
  \begin{aligned}
    \mbox{s.t.\ for all i,j}&& \label{eq2} \\
    & \mathbf w^{\top}\mathbf x_i^j- b_j \le  -1+\chi_i^j  \\
    &\mathbf w^{\top}\mathbf x_i^{j+1} - b_j\ge +1-\xi_i^{j+1} \\
    &b_j\le b_{j+1} \\ 
    &\chi_i^j\ge 0, \xi_i^j\ge 0 
  \end{aligned}
\end{equation}

where $\chi_i^j$ and $\xi_i^j$ are slack variables, and the thresholds $b_j$ for $j=1,\ldots, l-1$ determine the boundaries which separate  the classes, $b_j$ referring to the boundary in between bin $j$ and bin $j+1$.
The hyper-parameter $C>0$ controls the trade-off of the margin and number of errors and it can be chosen through cross validation.
We adapt the problem from~\cite{svmo}, which uses $L_2$ regularization,
and use $L_1$ regularization in (Eq.~\ref{eq:svm_ordreg_exc}), aiming for sparse solutions.
In this definition the linear ordering of classes is enforced \emph{explicitly} through constraint $b_j\le b_{j+1}$.
When we refer to (\ref{eq2}) in the future, we specifically refer to the constraints of the problem.

\paragraph*{Implicit Order Constraints}
Another definition first highlighted in~\cite{chu_new_2005} enforces the ordering implicitly, by requiring that all data of bin $1$ to $j$ are embedded below the threshold $b_j$, all data from bins $j+1$ to $l$ are above the threshold.
This leads to the implicitly constrained problem:
\begin{equation}
  \label{eq:svm_ordreg_imp}
\begin{aligned}
    \min_{\mathbf w,\mathbf b,\bm \chi,\bm \xi}
    &~\hfill\frac{1}{2} \|\mathbf w\|_1 + C \sum_{j=1}^{l-1}
    \left( \sum_{k=1}^{j}\sum_{i=1}^{n^k} \chi_{ki}^j +
    \sum_{k=j+1}^{l}\sum_{i=1}^{n^k}\xi_{ki}^j
    \right)\\
    \mbox{subject to}&\\
    &\mathbf w^{\top}\mathbf x_i^k -b_j \le -1+\chi_{ki}^j, \quad \chi_{ki}^j \geq 0,\\ 
    &~\hfill\text{\quad  for } k=1,\dots,j \text{ and } i=1,\dots,m_k;\\
    &\mathbf w^{\top}\mathbf x_i^{k} - b_j \ge +1 - \xi_{ki}^j, \quad \xi_{ki}^j \geq 0\\ 
    &~\hfill\text{\quad  for } k=j+1,\dots,l \text{ and } i=1,\dots,m_k.
\end{aligned}
\end{equation}
Again, we adapt the existing problem from~\cite{chu_new_2005} and replace the existing regularization $\|\mathbf w\|_2$ with $\|\mathbf w\|_1$ to induce sparsity.
In this definition, not only neighbouring classes are contributing to the overall loss of in between boundaries, but all other classes, as well.
This can lead to more robust results in particular in the case of outliers, as shown in~\cite{chu_new_2005}, but higher computational demand.

In the following we introduce feature relevance bounds for the explicit variant which is an extension from existing work for simple linear classification in~\cite{christina}.
The definition for the implicit variant is very similar and can be found in~\ref{apx:rele_bounds_implicit}.

\section{Feature Relevance Bounds for Ordinal Regression with Explicit Order}\label{sec:relev_bounds}
Assume a training set $X$ is given.
We denote an optimum solution of problem~(\ref{eq:svm_ordreg_exc}) as $(\tilde{\mathbf w},\tilde{\mathbf b},\tilde{\bm{\xi}}, \tilde{\bm{\chi}})$. 
This solution induces the value
\[\mu_X:=\frac{1}{2}\|\tilde{\mathbf w}\|_1 +C\cdot \sum_{i,j} \left(\tilde \chi_i^j+\tilde \xi_i^j\right)\] which is uniquely determined by $X$.
The quantity $\mu_X$ is unique by definition, albeit the solution ($\tilde{\mathbf w},\tilde{\mathbf b},\tilde{\bm{\xi}}, \tilde{\bm{\chi}}$)  is not.

We are interested  in the {class of equivalent good hypotheses}, i.e.\ all weight vectors $\mathbf{w}$ which yield (almost) the same quality as regards the regression error and generalization ability as the function induced by $\tilde{\mathbf{w}}$.
This class might contain an infinite number of alternative hypothesis: in the context of correlated features, for example, we can trade one feature for the other. However, the function class cannot explicitly be computed,
since the generalization ability is unknown for future data.
We use the following surrogate induced by $\mu_X$
\begin{eqnarray}\notag
  F_{\delta}(X)&:=&\{\mathbf w\in\mathbb{R}^n\:|\:\exists \mathbf b,\bm{\xi}, \bm{\chi} \mbox{ such that constraints (\ref{eq2}) hold,}\\\label{eq4}
  && \frac{1}{2}\|\mathbf w\|_1+ C\cdot
  \sum_{i,j}\left(\xi_i^j+\chi_i^j\right) \le (1+\delta)\cdot \mu_X\}
\end{eqnarray}

These constraints ensure the following properties:
\begin{enumerate}
  \item
        The empirical error of equivalent functions in $F_{\delta}(X)$ is minimum, as measured by the slack variables.
  \item The loss of the generalization ability  is limited, as guaranteed by a small $L_1$-norm of the
        weight vector and learning theoretical guarantees as provided, e.g.\ by Theorem 7 in~\cite{marginbounds} and Corollary 5 in~\cite{zhang}.
\end{enumerate}
The parameter $\delta\ge 0$ quantifies the tolerated deviation to accept a function as yet good enough, C is determined by Problem (\ref{eq:svm_ordreg_exc}).

Solutions $\mathbf{w}$ in $F_{\delta}(X)$ are sparse in the sense that irrelevant features are uniformly weighted as $0$ for all solutions in $F_{\delta}(X)$. Relevant but potentially redundant features can be weighted arbitrarily, disregarding sparsity, similar in spirit to
the ElasticNet; yet the latter weights mutually redundant  features equally and can therefore hide the relevance in the case of many redundant features~\cite{zou2005}.
In this contribution we are interested in the relevance of features for forming good hypotheses; more precisely, we are interested in the following more specific characteristics:

\begin{itemize}
  \item {\bf Strong relevance} of feature $\feature$ for $F_{\delta}(X)$: Is feature $\feature$ relevant for all hypotheses in $F_{\delta}(X)$, i.e.\ all
        weight vectors $\mathbf w\in F_{\delta}(X)$ yield $w_{\feature}\not= 0$?
  \item {\bf Weak relevance} of feature $\feature$ for $F_{\delta}(X)$: Is feature $\feature$ relevant for at least one hypothesis in $F_{\delta}(X)$ in the sense that one
        weight vector $\mathbf w\in F_{\delta}(X)$ exists with $w_{\feature}\not= 0$, but this does not hold for all weight vectors in $F_{\delta}(X)$?
  \item {\bf Irrelevance} of feature $\feature$ for $F_{\delta}(X)$: Is feature $\feature$ irrelevant for every hypothesis in $F_{\delta}(X)$, i.e.\ all
        weight vectors $\mathbf w\in F_{\delta}(X)$ yield $w_{\feature}= 0$?
\end{itemize}

A feature is irrelevant for $F_{\delta}(X)$ if it is neither strongly nor weakly relevant.
The questions of strong and weak relevance can be answered via the following optimization problems:

\begin{description}
  \item[Problem $\mathbf{minrel}(\feature)$:]
        \begin{eqnarray}
          \min_{\mathbf w,\mathbf b,\bm \chi, \bm \xi} && |w_{\feature}| \label{eq5}\\\notag
          \mbox{s.t.\ for all } i,j&& \mbox{conditions (\ref{eq2}) hold and}\\
          &&\frac{1}{2}\|\mathbf{w}\|_1 + C\cdot \sum_{k,l}\left(\chi_k^l+\xi_k^l \right)\le (1+\delta)\cdot\mu_X \label{eq6}
        \end{eqnarray}
        Here $|w_{\feature}|$ denotes the absolute value of feature $\feature$ in $w$.
        Feature $\feature$ is strongly relevant for $F_{\delta}(X)$ iff $\mathrm{minrel}(\feature)$ yields an optimum larger than $0$.

  \item[Problem $\mathbf{maxrel}(\feature)$:]
        \begin{eqnarray}
          \max_{\mathbf w,\mathbf b,\bm \chi, \bm \xi} && |w_{\feature}| \label{eq8}\\\notag
          \mbox{s.t.\ for all } i,j&& \mbox{conditions (\ref{eq2}) and (\ref{eq6}) hold}
        \end{eqnarray}
        Feature $\feature$ is weakly relevant for $F_{\delta}(X)$ iff $\mathrm{minrel}(\feature)$ yields an optimum at $0$
        and $\mathrm{maxrel}(\feature)$ yields an optimum larger than $0$.
\end{description}

These two optimization problems span a real-valued interval for every feature $\feature$ with the result of
$\mathrm{minrel}(\feature)$ as lower and $\mathrm{maxrel}(\feature)$ as upper bound.
This interval characterizes the range
of weights for $\feature$ occupied by good solutions in $F_{\delta}(X)$. Hence, besides information about a feature's relevance,
some indication about the degree up to which a feature is relevant or can be substituted by others, is given.
Note, however, that the solutions are in general not consistent estimators of an underlying \lq true\rq\ weight vector as regards its exact value,
as has been discussed, e.g.\ for lasso~\cite{consistencylasso}.
For consistency, it is advisable to use L$_2$ regularization after the selection of a set of relevant features.

\subsection{Generalization Bounds}
In the beginning of Section~\ref{sec:relev_bounds} we introduced the set $F_\delta(X)$ of all equivalent good hypotheses which yield (almost) the same quality regarding regression error and generalization ability. However, the impact of the norm of $\mathbf{w}$ and the high loss $\sum_{i,j} \left( \tilde{\chi}_i^j + \tilde{\xi}_i^j \right)$ are not considered separately, i.e.\ a low norm of $\mathbf{w}$ allows a high loss, and vice versa. We would like to control the generalization error by means of $l_1$-regularization. To do so, we consider both quantities separately, i.e.\  we define
\begin{eqnarray}\notag
  \mathcal{H}_\delta(\tilde{\mathbf w}) &:=&\{\mathbf w\in\mathbb{R}^n\:|\:\exists \mathbf b,\bm{\xi}, \bm{\chi} \mbox{ such that constraints (\ref{eq2}) hold,}\\
  && \|\mathbf w\|_1 \leq (1+\delta) \| \tilde{\mathbf{w}}  \|_1 \text{ and } \\
  && \left. \sum_{i,j}\left(\xi_i^j+\chi_i^j\right) \leq \sum_{i,j}\left(\tilde{\xi_i}^j+\tilde{\chi}_i^j\right) \right\rbrace.
\end{eqnarray}
This allows us to extend the results from~\cite{christina} to our scenario, i.e.\  show that the generalization error of all hypothesis with the same or a lower high loss is bounded by means of the $l_1$-regularization. Recall Theorem 26.15 from Understanding Machine Learning~\cite{gen}:

\begin{theorem}\label{thm0}
\newcommand{\R}{\mathbb{R}}
Suppose that $\mathcal{D}$ is a distribution on $X \times Y$ such that with probability 1 we have $\Vert x \Vert_\infty \leq R$. Let $\mathcal{H} = \{ \mathbf{w} \in \R^d \mid \Vert \mathbf{w} \Vert_1 \leq B \}$ and let $l : \mathcal{H} \times X \times Y \to \R$ be of the form $l(\mathbf{w},(x,y)) = \phi(\langle \mathbf{w}, x \rangle, y)$ where $\phi : \R \times Y \to \R$ is such that for all $y \in Y$, the function $a \mapsto \phi(a,y)$ is $\eta$-Lipschitz and such that $\max_{a \in [-RB,RB]} |\phi(a,y)| \leq c$. Then, for any $\tau \in (0,1)$ with probability of at least $1-\tau$ over the choice of i.i.d. sample of size $n$, for all $\mathbf{w} \in \mathcal{H}$,
\begin{align*}
    \mathbb{E}_{(x,y) \sim \mathcal{D}} [l(\mathbf{w},x,y)] \leq \frac{1}{n} \sum_{i=1}^n l(\mathbf{w},x_i,y_i) + 2 \eta RB \sqrt{\frac{2 \log(2d)}{n}} + c \sqrt{\frac{2 \ln(2/\tau)}{n}}.
\end{align*}
\end{theorem}

To apply this theorem we have to reformulate our classifier as a collection of binary classifiers. Since all classes use the same subspace spanned by $\mathbf{w}$ it is enough to distinguish neighbouring classes, i.e.\  every $b_j$ gives rise to a classifier that allows us to decide whenever $x$ belongs to one of $0,\dots,j$ or $j+1, \dots ,|L|$. 
Consider the ramp loss
\begin{align*}
    l_{\prec j} (\mathbf{w},\mathbf{b},x,y) &= \min\{1,\max\{0,1-\mathbf{1}_{y \prec j} (\mathbf{w}^{\top} x - b_j)\}\}, \\
    l_j(\mathbf{w},\mathbf{b},x,y) &= l_{\leq j}(\mathbf{w},\mathbf{b},x,y)+l_{\geq j}(\mathbf{w},\mathbf{b},x,y), \\
    l(\mathbf{w},\mathbf{b},x,y) &= l_y(\mathbf{w},\mathbf{b},x,y)
\end{align*}
where $\mathbf{1}_{y \prec j} = 1$ if $y \prec j$ and $-1$ otherwise for some comparison operation $\cdot \prec \cdot$.
Notice that $l$ corresponds to the implicit order constrains, which is an upper bound for the explicit loss where only neighbouring classes are considered, rather than all classes.
By using this loss function it is clear that the 
loss of the original classifier is bounded by the sum of all those binary classifiers.
Since the ramp loss is $1$-Lipschitz and maps to the interval $[0,1]$ we may apply Theorem~\ref{thm0} to obtain
\begin{align*}
    \mathbb{E}_{(x,y) \sim \mathcal{D}} [l(\mathbf{w},x,y)] 
    &\leq \mathbb{E}_{(x,y) \sim \mathcal{D}} \left[ \sum_{j = 1}^{|L|} (l_{\leq j}(\mathbf{w},x,y) +  l_{\geq j}(\mathbf{w},x,y)) \right]
    \\&
    = \sum_{j = 1}^{|L|} \left( \mathbb{E}_{(x,y) \sim \mathcal{D}} \left[ l_{\leq j}(\mathbf{w},x,y) \right] + \mathbb{E}_{(x,y) \sim \mathcal{D}} \left[ l_{\geq j}(\mathbf{w},x,y) \right] \right)
    \\&\leq \sum_{j = 1}^{|L|} \left(\frac{1}{n} \sum_{i=1}^n (l_{\leq j}(\mathbf{w},x_i,y_i) + l_{\geq j}(\mathbf{w},x_i,y_i)) \right. \\&\qquad\qquad \left. + 4 RB \sqrt{\frac{2 \log(2d)}{n}} + 2\sqrt{\frac{2 \ln(2/\tau)}{n}} \right)
\end{align*}
for all $\mathbf{w}$ such that $\Vert \mathbf{w} \Vert_1 \leq B$ with probability $1-\tau$ over the choice of sample. In particular, setting $\rho_j = \sum_i \tilde{\xi}_i^j+\tilde{\chi}_i^j$ and $\rho = \sum_j \rho_j$ to the hinge loss of the baseline classifier and using the fact that the hinge loss upper bounds ramp loss, this gives rise to
\begin{align*}
    L_{\mathcal{D}} (\tilde{\mathbf{w}},\tilde{\mathbf{b}}) &\leq |L|\left(\frac{\rho}{n} + 4 \Vert \tilde{\mathbf{w}} \Vert_1 R \sqrt{\frac{2 \log(2d)}{n}} +  2\sqrt{\frac{2 \ln(2/\tau)}{n}}\right)
\end{align*}
for the generalization error of the baseline linear classifier $(\tilde{\mathbf{w}},\tilde{\mathbf{b}})$ and
\begin{align*}
    L_{\mathcal{D}} (h) &\leq |L|\left(\frac{\rho}{n} + 4 (1+\delta) \Vert \tilde{\mathbf{w}} \Vert_1 R \sqrt{\frac{2 \log(2d)}{n}} + 2 \sqrt{\frac{2 \ln(2/\tau)}{n}}\right)
\end{align*}
for all $h \in \mathcal{H}_\delta(\tilde{\mathbf w})$, with probability at least $1-\tau$ over the choice of training sample, i.e.\  our choice of constraints allow the generalization error upper bound to increase by $4 \delta \Vert \tilde{\mathbf{w}} \Vert_1 |L| R \sqrt{\frac{2 \log(2d)}{n}}$.

\subsection{Feature Relevance Bounds as Linear Problem}\label{LP_relev_bounds}
The problems from Section~\ref{sec:relev_bounds} are not yet linear problems, but they can be transferred to linear optimization problems, for which particularly efficient solvers are available.
\begin{theorem}\label{thm1}
  Problem $\mathrm{minrel}(\feature)$ is equivalent to the following linear optimization problem:
  \begin{eqnarray}\notag
    {\mathbf{minrel}^*(\feature):}\min_{\mathbf w, \mathbf w,\mathbf b,\bm \chi, \bm \xi} && \hat w_{\feature}\\\notag
    \mbox{s.t.\ for all } i,j && \mbox{conditions (\ref{eq2}) hold}\\
    &&\frac{1}{2}\sum_k\hat w_k + C\cdot \sum_{k,l}\left(\chi_k^l+\xi_k^l \right)\le (1+\delta)\cdot\mu_X\\
    &&  w_i \le \hat w_i, \, -w_i \le  \hat w_i \label{eq9}
  \end{eqnarray}
  Problem $\mathrm{maxrel}(\feature)$ can be solved by taking the optimum of the following  two linear optimization problems:
  \begin{eqnarray}\notag
    \mathbf{maxrel}^*_{\mathrm{pos}}(\feature):\max_{\mathbf w, \mathbf w,\mathbf b,\bm \chi, \bm \xi} && \hat w_{\feature}\\\notag
    \mbox{s.t.\ for all } i,j && \mbox{conditions (\ref{eq2}) hold}\\\notag
    &&\frac{1}{2}\sum_k\hat w_k + C\cdot \sum_{k,l}\left(\chi_k^l+\xi_k^l \right)\le (1+\delta)\cdot\mu_X \\\notag
    &&  w_i \le \hat w_i, \, -w_i \le  \hat w_i \\
    &&\hat w_{\feature}\le w_{\feature}\label{eq11}
  \end{eqnarray}
  and the problem
  \begin{eqnarray}\notag
    \mathbf{maxrel}^*_{\mathrm{neg}}(\feature):\max_{\mathbf w, \mathbf w,\mathbf b,\bm \chi, \bm \xi} && \hat w_{\feature}\\\notag
    \mbox{s.t.\ for all } i,j && \mbox{conditions (\ref{eq2}) hold}\\\notag
    &&\frac{1}{2}\sum_k\hat w_k + C\cdot \sum_{k,l}\left(\chi_k^l+\xi_k^l \right)\le (1+\delta)\cdot\mu_X \\\notag
    &&  w_i \le \hat w_i, \, -w_i \le  \hat w_i \\
    &&\hat w_{\feature}\le - w_{\feature}\label{eq12}
  \end{eqnarray}
\end{theorem}
The proof can be found in the appendix.
\\\
In practice, it might be a good strategy to split constraint (\ref{eq5}) into two, separately limiting
the weight vector
\[\frac{1}{2}\sum_k\hat w_k \le(1+\delta)\cdot \|\tilde{\mathbf w}\|_1\]
and error term \[\sum_{k,l}\left(\chi_k^l+\xi_k^l\right)\le\sum_{k,l}\left(\tilde\chi_k^l+\tilde\xi_k^l\right)\]
where the symbols marked $\tilde{\cdot}$ refer to the optimum solution of the original margin-based ordinal regression problem.
This split enables us to better control the loss of generalization ability and error terms, and it also mediates the
dependency on the hyper-parameter $C$ of the space of equivalent good functions. At a small down-side, this split
depends on the found solution and it is no longer uniquely defined by the given training data, albeit we did not observe large variation in practical applications.

\section{Learning using Privileged Information}\label{sec:lupi}
Let us shortly recall the classical setting considered so far:
Given ordered class labels $L=\{1,2,\ldots,l\}$ and training data $X=\{\mathbf x_i^j\in\mathbb{R}^n\:|\: i=1,\ldots,m_j,\, j\in L\}$ where data point $x_i^j$ is assigned the class label $j\in L$.
The full data set has size $m:=m_1+\dots+m_l$.
Here the index $j$ refers to the ordinal target variable (represented by $b_j$) the  data point $x_i^j$  belongs to.

In the LUPI setting, we work with two types of information $X$ and $\Xp = \{{\xp}^j \in \IR^{n^{*}}\:|\:i=1,\ldots, m_j, j\in L\} $ which is a set of additional information commonly called privileged information (PI) where $p$ is the amount of privileged features we have available.
The information is privileged in the sense that it is not available in the testing and prediction phase, and it is only present when training the model.
This fact does not necessarily imply that the privileged information is of higher quality or exhibits correlation with the label $y$ at all. Rather, there are reasons why it cannot be gathered at prediction time: examples are too costly computations (such as extensive feature preprocessing), unavailability of sensors, unavailability of the information (such as information which is available only in retrospective, or privacy issues which prevent gathering the data (such as personal information).
$X$ and $\Xp$, in general, do not have to share the same space or modality.
As an example, $X$ could cover numerical features, and $\Xp$ could be textual input from an expert.

\subsection{Modelling Slacks in Ordinal Regression}\label{sub:modelling_slacks_in_ordinal_regression}
There are several ways to integrate privileged information into the learning model~\cite{lopez-paz2015}.
In the following we only consider \emph{similarity control} where privileged information is interpreted as the teacher giving hints about the difficulty for each training example.
These hints can be incorporated into an SVM by means of slack variables which was shown in~\cite{tang} already.
In the following we will extend our \emph{explicit} definition of ordinal regression  to handle privileged information by adapting similarity control as used in~\cite{tang}.

We recall that in the explicit variant two types of slacks are used.
Each slack value represents a deviation from the classification rule.
In the LUPI case, we replace $\chi_i^j$ by
\[
  \fprivchi:= \privfuncchi
\]
and $\xi_i^j$ by the function
\[
  \fprivxi:= \privfuncxi.
\]

\begin{eqnarray}
  \label{eq:lupi_svm_ordreg}
  \notag\min_{\mathbf w,\mathbf b, \omegap[],\mathbf{\dpriv[]}}
  & & \svmpriv\\
  & & \mbox{s.t.\ for every }j=1,\dots,l-1,\\\notag
  & & \mathbf w^{\top}\mathbf x_i^j- b_j \le  -1+\fprivchi \\\notag
  & & \mathbf w^{\top}\mathbf x_i^{j+1} - b_j\ge +1-\fprivxi[j+1]\\\notag
  & & b_j\le b_{j+1} \\ \notag
  & & \fprivchi \ge 0, \fprivxi \ge 0 \notag
\end{eqnarray}

$\gamma$ is an additional hyperparameter to scale the influence of privileged information.
This allows us to reject nonsense PI by simplifying the model and relying solely on $X$ when considering a cross validation scheme where we expect better generalization ability by a simpler model.
The adaption of~\cite{tang} now enables us to define relevance bounds as in Section~\ref{sec:relev_bounds}.

\subsection{Feature Relevance Bounds for Ordinal Regression with Privileged Information}\label{sub:feature_relevance_bounds_for_ordinal_regression_with_pi}

We now consider two sets of features.
In the following we define bounds for both regarding their relevance to the machine learning procedure when both sets are present.
Because PI are not present while predicting they are always irrelevant for that phase. They are relevant to
speed up learning by mediating the distribution of slack variables.

Assume a training set  $X=\{\mathbf x_i^j\in\mathbb{R}^n\}$ and $\Xp~=~\{{\xp}^j \in \IR^{n^{*}}\}$.
Further we define \[
  \privlossfunc:=\privloss
\] as the total slack loss of problem~(\ref{eq:lupi_svm_ordreg}).
Denote an optimum solution of the problem as
$(\tilde{\mathbf w},\tilde{\mathbf{b}},\tilde{\mathbf{w}}^{*}_{\chi},\tilde{\mathbf{w}}^{*}_{\xi}, \tilde{\dpriv[\chi]}, \tilde{\dpriv[\xi]} )$
and its total loss as $\tilde{\privlossfunc}$.
Analogous to Section~\ref{sec:relev_bounds}, this solution induces the
value
\[
  \privmu:= \optsvmpriv.
\]

Furthermore, we use the following proxy induced by $\privmu$
\begin{eqnarray}
  \label{eq:lupi_proxy}
  \notag
  \privproxy&:=&
  \notag
  \{\mathbf w\in \IR^n,\ \omegap[\chi],\omegap[\xi] \in \IR^{n^*}\ \:|\: \exists \mathbf b, d_{\chi},d_{\xi}\\
  &&\mbox{ such that constraints (\ref{eq:lupi_svm_ordreg}) hold and}\\
  \notag
  && \wnorm + \wprivnorm + \privlossfunc \le (1+\delta)\cdot \privmu\}
\end{eqnarray}

This proxy allows us to define similar feature relevances as found in Section~\ref{sec:relev_bounds} for non-privileged feature $\feature$ in  $X$:
\begin{itemize}
  \item {\bf Strong relevance} of feature $\feature$ for $\privproxy$: Is feature $\feature$ relevant for all hypotheses in $\privproxy$, i.e.\ all
        weight vectors $\mathbf w\in \privproxy$ yield $w_{\feature}\not= 0$?
  \item {\bf Weak relevance} of feature $\feature$ for $\privproxy$: Is feature $\feature$ relevant for at least one hypothesis in $\privproxy$ in the sense that one
        weight vector $\mathbf w\in \privproxy$ exists with $w_{\feature}\not= 0$, but this does not hold for all weight vectors in $\privproxy$?
  \item {\bf Irrelevance} of feature $\feature$ for $\privproxy$: Is feature $\feature$ irrelevant for every hypothesis in $\privproxy$, i.e.\ all
        weight vectors $\mathbf w\in \privproxy$ yield $w_{\feature}= 0$?
\end{itemize}

and similarly for feature $\pfeature$ in $\Xp$ with $\omegap[\bullet] := \{\omegap[\chi],\omegap[\xi]\ \:|\: (\omegap[],\omegap[\chi],\omegap[\xi]) \in \privproxy\}$:

\begin{itemize}
  \item {\bf Strong relevance} of feature $\pfeature$ for $\privproxy$: Is feature $\pfeature$ relevant for all hypotheses in $\privproxy$, i.e.\  for all $\omegap[\bullet]$ in $\privproxy$ at least one weight vector in $\omegap[\bullet]$ for one bin of the ordered classes  yields $w^{*}_{\bullet \pfeature}\not= 0$?
  \item {\bf Weak relevance} of feature $\pfeature$ for $\privproxy$: Is feature $\pfeature$ relevant for at least one hypothesis in $\privproxy$ in the sense that one
        weight vector $\omegap[\bullet]$ exists with $w^{*}_{\bullet \pfeature}\not= 0$, but this does not hold for all $\omegap[\bullet]$ in $\privproxy$?
  \item {\bf Irrelevance} of feature $\pfeature$ for $\privproxy$: Is feature $\pfeature$ irrelevant for every hypothesis in $\privproxy$, i.e.\ all
        weight vectors $\omegap[\bullet]$ yield $w^{*}_{\bullet \pfeature}= 0$?
\end{itemize}
A feature is irrelevant for $\privproxy$ if it is neither strongly nor weakly relevant.

The questions of strong and weak relevance can be answered via the following optimization problems:

\begin{description}
  \item[Problem $\mathbf{minrel}(\pfeature)$:]
        \begin{eqnarray}
          \max_{\bullet \in \{\chi,\xi\}} \min_{\mathbf w,\omegap[\bullet] ,\mathbf b,\dpriv[\bullet
            ]}                        & & |w^{*}_{\bullet \pfeature}| \label{eq:minrel_lupi}\\\notag
          \mbox{s.t.\ for all } i,j & & \mbox{conditions (\ref{eq:lupi_svm_ordreg}) hold and}\\
          & & \wnorm + \wprivnorm + \privlossfunc \le (1+\delta)\cdot\privmu \label{eq:minrel_lupi_constraint}\notag
        \end{eqnarray}
        Because of two slack functions and the corresponding weights  $\mathbf{w}^{*}_{\chi}$ and $\mathbf{w}^{*}_{\xi}$ we need to optimize two inner feature relevancies $|w^{*}_{\bullet \pfeature}|$.
        To aggregate them to a global feature relevance we take the maximum to express that a feature could be used only in one of both functions, i.e.\  it is not relevant for all slack functions but at least in one.
        One could define an additional relevance classification by taking into account cases where the $\min\min >0$, i.e.\  the feature is relevant for all slack functions.
        In the following we limit ourselves to the former case.

        Feature $\pfeature$ is strongly relevant for $\privproxy$ iff $\mathrm{minrel}(\pfeature)$ yields an optimum larger than $0$.

  \item[Problem $\mathbf{maxrel}(\pfeature)$:]
        \begin{eqnarray}
          \max_{\bullet \in \{\chi,\xi\}} \max_{\mathbf w,\omegap[\bullet] ,\mathbf b,\bm \chi, \bm \xi} & & |w^{*}_{\bullet \pfeature}| \label{eq:maxrel_lupi}\\\notag
          \mbox{s.t.\ for all } i,j            & & \mbox{conditions (\ref{eq:lupi_svm_ordreg}) hold and}\\
          & & \wnorm + \wprivnorm + \privlossfunc \le (1+\delta)\cdot\privmu \label{eq:maxrel_lupi_constraint}\notag
        \end{eqnarray}
        Similar to the first problem we consider the maximum inner feature relevance to express the global feature relevance.

        Feature $\pfeature$ is weakly relevant for $\privproxy$ iff $\mathrm{minrel}(\pfeature)$ yields an optimum $0$
        and $\mathrm{maxrel}(\pfeature)$ yields an optimum larger than $0$
\end{description}

\subsection{Privileged Feature Relevance Bounds as Linear Problem}\label{sub:lupi_lp}
Both problems can be transferred to linear optimization problems:
\begin{theorem}\label{thm2}
  Problem $\mathrm{minrel}(\pfeature)$ is equivalent to taking the maximum over following two linear optimization problems:

  \begin{eqnarray}
    \notag
    \mathbf{minrel}^*_{\mathrm{\chi}}(\pfeature):\\
    \min_{ \substack{\mathbf w,\hat{\mathbf w}, \omegap[\chi],
     \widehat{\omegap[\chi]}, \omegap[\xi], \widehat{\omegap[\xi]},\\
      \mathbf b,\dpriv[\chi], \dpriv[\xi]}}
    & & \wph[\chi \pfeature] \label{eq:minrel_lp_lupi_chi}\\\notag
    \mbox{s.t.\ for all } i,j & & \mbox{conditions (\ref{eq:lupi_svm_ordreg}) hold and}\\\notag
    &&\frac{1}{2}\lonesum[\hat w]_k + \frac{\gamma}{2}\lonesum[\hat{w}^*_{\chi k}] + \frac{\gamma}{2}\lonesum[\hat{w}^*_{\xi k}] \ +  \privlossfunc \le (1+\delta)\cdot\mu_X \\\notag
    &&  \loneconstraint{w}{i}\\\notag
    &&  \loneconstraint{\chi}{i}\\\notag
    &&  \loneconstraint{\xi}{i}\\\notag
  \end{eqnarray}
  and
  \begin{eqnarray}
    \notag
    \mathbf{minrel}^*_{\mathrm{\xi}}(\pfeature):\\
    \min_{ \substack{\mathbf w,\hat{\mathbf w}, \omegap[\chi], \widehat{\omegap[\chi]}, \omegap[\xi], \widehat{\omegap[\xi]},\\ \mathbf b,\dpriv[\chi], \dpriv[\xi]}}
    & & \wph[\xi \pfeature] \label{eq:minrel_lp_lupi_xi}\\\notag
    \mbox{s.t.\ for all } i,j & & \mbox{conditions (\ref{eq:lupi_svm_ordreg}) hold and}\\\notag
    & & \frac{1}{2}\lonesum[\hat w]_k + \frac{\gamma}{2}\lonesum[\hat{w}^*_{\chi k}] + \frac{\gamma}{2}\lonesum[\hat{w}^*_{\xi k}] \ +  \privlossfunc \le (1+\delta)\cdot\mu_X \\\notag
    & & \loneconstraint{w}{i}\\\notag
    & & \loneconstraint{\chi}{i}\\\notag
    & & \loneconstraint{\xi}{i}\\\notag
  \end{eqnarray}

  For $\mathrm{maxrel}(\pfeature)$ we first define the linear optimization problem

  \begin{eqnarray}
    \notag
    \mathbf{maxrel}^*_{\mathrm{\lambda,\bullet}}(\pfeature):\\
    \max_{ \substack{\mathbf w,\hat{\mathbf w}, \omegap[\chi], \widehat{\omegap[\chi]}, \omegap[\xi], \widehat{\omegap[\xi]},\\ \mathbf b,\dpriv[\chi], \dpriv[\xi]}}
    & & \wph[\bullet \pfeature] \label{eq:minrel_lp_lupi_pos}\\\notag
    \mbox{s.t.\ for all } i,j & & \mbox{conditions (\ref{eq:lupi_svm_ordreg}) hold and}\\\notag
    & & \frac{1}{2}\lonesum[\hat w]_k + \frac{\gamma}{2}\lonesum[\hat{w}^*_{\chi k}] + \frac{\gamma}{2}\lonesum[\hat{w}^*_{\xi k}] \ +  \privlossfunc \le (1+\delta)\cdot\mu_X \\\notag
    & & \loneconstraint{w}{i}\\\notag
    & & \loneconstraint{\chi}{i}\\\notag
    & & \loneconstraint{\xi}{i}\\\notag
    &&\hat w^*_{\bullet \pfeature} \le \lambda \cdot w^*_{\bullet \pfeature}
  \end{eqnarray}
  such that
  \[
    \mathrm{maxrel}(\pfeature) := \max_{\substack{\lambda \in \{-1, +1\},\\ \bullet \in \{\chi, \xi \}}} \mathrm{maxrel}^*_{\mathrm{\lambda,\bullet}}(\pfeature),
  \] i.e.\  the maximum of four linear problems.
\end{theorem}
A proof of this theorem is similar to Section~\ref{LP_relev_bounds} and is omitted for the sake of brevity.

\section{Relevance Bounds for Feature Selection}\label{sub:handling_numerical_instabilities}
While the relevance bounds should give truthful indication of feature relevance, in practice the discrimination between relevant and irrelevant features is challenging:
variations of the underlying distributions of the features have the implication that thresholds for feature relevance can vary for different features.
The use of slack variables in the overall model and thus our relevance bounds allow variation in the contribution of features which improves finding stable solutions but also adds noise.
This is exacerbated by the behaviour of linear programming solvers, which often have exhibit loss of precision.
For relevance bounds specifically, even if feature $\feature$ is independent we often observe $\text{maxRel}(\feature)>0$ and $0 < \text{minRel}(\feature)<10^{-5}$.

We do not aim for a data independent threshold to discriminate between noise and relevant features.
Instead, we introduce distribution dependent thresholds: we estimate the distribution of relevances of noise features given the model constraints.
We expect for a given model class defined by $F_{\delta}(X)$ the same amount of slackness in the relevances for irrelevant variables.
This slackness is introduced by the parameters of the algorithm itself ($\delta$, $\regparameter$) and the LP-solvers internal ones and should be similar for truly non-correlated variables.
Therefore, we propose to estimate the parameters of a normal distribution and the corresponding prediction interval $\predi$ to obtain a data dependent threshold~\cite{geisser1993}.
An existing work proposes a similar resampling based approach to estimate a stopping threshold for a forward feature selection approach~\cite{francois2007}.

To estimate this noise distribution we use randomly permuted input features from $X$ to imitate irrelevant features.
We define $p(\feature)$ as the random permutation of values in $\feature$ and $\datawithperm := \{X\setminus \feature\} \cup \perm \feature$ as the dataset where $\feature$ was replaced by its random permutation.
With these we define two random sample populations
\[
  \probedist{\maxrel}:=\{\maxrel(\perm \feature, \datawithperm)\ |\ \text{where } \feature \text{ randomly chosen from } X\}
\]
and
\[
  \probedist{\minrel}:=\{\minrel(\perm \feature, \datawithperm)\ |\ \text{where } \feature \text{ randomly chosen from } X\}
\]
where a population with $n$ samples is defined as $\probedist{\cdot}_n$.

The prediction interval is then defined as
\[
  \PI{\cdot}.
\]
Here $\overline {\pi}_{n}$ denotes the sample mean and $\sigma(x)$ the standard deviation, and $T$ represents Student's t-distribution with $n-1$ degrees of freedom.
The size of $\predi$ depends on parameter $p$, the expected probability that a new value is included in the interval.
We propose default values of $p=0.999$ for a low false positive rate and $n\geq50$ which yielded robust thresholds for common feature set sizes in our experiments without adding too many computations to the complexity, which we analyse in Section~\ref{sub:complexity}.

To classify feature $\feature$ as irrelevant we check if its relevance bounds are element of our prediction intervals.
We therefore replace the theoretical classifications from Section~\ref{sec:relev_bounds} with the following:
\begin{itemize}
  \item {\bf Strong relevance}: $\maxrel(\feature) \notin \predi(\maxrel) \land \minrel(\feature) \notin \predi(\minrel)$
  \item {\bf Weak relevance}: $\maxrel(\feature) \notin \predi(\maxrel) \land \minrel(\feature) \in \predi(\minrel)$
  \item {\bf Irrelevance}: $\maxrel(\feature) \in \predi(\maxrel) \land \minrel(\feature) \in \predi(\minrel)$
\end{itemize}

\subsection{Time complexity}\label{sub:complexity}
In the following we outline the scaling behaviour of our proposed method for feature selection.
Our method can be divided in three separate computational steps which differ in their algorithmic complexity.
We consider a problem with $n$ samples and $d$ features.

The initial baseline solution is analogue to a standard ordinal regression SVM solution which can be solved using the sequential minimal optimization (SMO) algorithm~\cite{platt1998,svmo} which is in \bigO{n^3}.
The relevance bounds are given by a set of linear programs for which interior point methods exist~\cite{karmarkar1984,vaidya1989,cohen2018} which are in \bigO{n^{2.5}}.
This complexity bound is very general and one could reformulate and adapt these problems using existing outlines~\cite{joachims2006,hsieh2008}.
In the normal setting we consider the constant $z=3$ for the number of linear programs needed (Section~\ref{LP_relev_bounds}) and $z=6$ in the LUPI setting (Section~\ref{sub:lupi_lp}) such that the relevance interval for each feature is in \bigO{zn^{2.5}}.
This results in \bigO{dzn^{2.5}} for all relevance bounds.
Additionally, we employ a permutation test approach which adds a constant $c$ additional LPs to achieve statistical stability which is overall in \bigO{cn^{2.5}}.
Overall our method is in \bigO{n^3 + (dz+c) n^{2.5}} when considering $n>d$.

Because the $dz+ c$ LPs are a significant factor, we proposed to solve them in parallel~\cite{pfannschmidt2019a} which we evaluate in~\ref{sec:scaling_eval}.

\section{Experiments}\label{sec:experiments}

We evaluate our methodology in two steps. 
First, we focus on our ordinal regression approach in the classical machine learning setting - using regular data.
Then we examine the adaption of our method to the LUPI paradigm - using data that incorporates privileged information.

\subsection{Classical Setting of ORP}\label{sub:classical_machine_learning}
In this section, we focus on our ordinal regression method for regular data.
We show the quality of our feature selection by evaluating the results of both the explicit and the implicit variant of our method, on theoretically generated data with know ground truth.
In addition, we compare both variants with regard to their classification accuracy and run time on standard benchmark datasets. 
The accuracy is measured using the \textit{Macro-averaged Mean Absolute Error (MMAE)} which is specifically designed for ordinal regression data with imbalanced classes:

\begin{equation}
  \label{eq:mmae}
  MMAE = \frac{1}{l}\sum_{j = 1}^{l}\frac{\sum_{i=1}^{m_j} \left | j - f(x_i^j) \right |}{m_j},
\end{equation}

where $l$ is the number of bins, $f$ refers to the bin the sample $x_i^j$ is assigned to by the learned model, and $m_j$ refers to the number of samples in class $j$.

The section is rounded off by an analysis of a real world data set, showcasing the insights that can be gained from our method.

\subsubsection{Artificial Data}\label{subs:artif_data}

We adapt the generation method presented in~\cite{christina} for ordinal regression.
By using equal frequency binning we convert the continuous regression variable into an ordered discrete target variable with five ordinal classes.
The data is generated from a suitable set of informative features. From those we form strongly relevant features by simply picking the desired number out of the informative set.
Weakly relevant features are created as linear combinations of informative features.
Finally, irrelevant features are drawn from random Gaussian noise.
All features are normalized to zero mean and unit variance.
The exact characteristics of the datasets used in our experiments are shown in Table~\ref{t:artificial_data}.

\begin{table}[H]
  \caption{Artificially created data sets with known ground truth.
    The model of which the data is drawn from is based on the strongly relevant features.
    The weakly relevant features are linear combinations of strong ones.
    Characteristics of the sets are taken from~\cite{christina} and~\cite{lukas}.
    All sets have target variables with five ordinal classes.}
  \label{t:artificial_data}
  \centering
  \begin{tabular*}{0.89\textwidth}{l c || c c c}
    \hline
    \hline
    \textbf{Dataset}           & \textbf{\#Instances} & \textbf{\#Strong} & \textbf{\#Weak} & \textbf{\#Irrelevant}  \\
    \hline
    Set 1 					& 150 & 6 & 0 & 6    	 \\
    Set 2					& 150 & 0 & 6 &	6		 \\
    Set 3				    & 150 & 3 & 4 &	3 		 \\
    Set 4					& 256 & 6 & 6 & 6 		 \\
    Set 5		    	 	& 512 & 1 & 2 & 11 		 \\
    Set 6					& 200 & 1 & 20 & 0		 \\
    Set 7					& 200 & 1 & 20 & 20      \\
    Set 8					& 1000 & 10 & 20 & 10      \\
    Set 9					& 1000 & 10 & 20 & 200      \\

    \hline
    \hline

  \end{tabular*}
\end{table}

For evaluation, we use the F-measure to quantify the detection of the all relevant feature set found by our method (dubbed feature relevance interval~-~FRI)\footnote{ Implementation in Python: \href{https://github.com/lpfann/fri}{https://github.com/lpfann/fri}} with regard to the true all relevant features of the data.

Because of the lack of other feature selection methods in this context we emulate the behaviour of lasso~\cite{tibshirani_regression_1996} and the ElasticNet (EN)~\cite{zou2005}.
For that we utilize a cross-validated recursive feature elimination\footnote{Implementation in Python: RFECV from scikit-learn}, using the ordinal regression model given by Equation~\ref{eq:svm_ordreg_exc} with an ElasticNet penalty and parameter $p$. 
The parameter $p$, controlling the ratio between the $L_1$ and $L_2$ norm of the EN model, is optimized with a search over the values $p \in \{0, 0.01, 0.1, 0.2, 0.5, 0.7,1\}$.
Setting $p=0$ corresponds to a lasso like sparsity constraint, and we test that scenario explicitly.
Our surrogates are called \svl[e] (lasso) and \en (EN), both based on the explicit variant.

Hyper parameters are selected according to 5-fold cross validation, and all scores are averaged over 30 independent runs. 

The results are given in
Table~\ref{t:toyperf_noise}, where $FRI_e$ and $FRI_i$ denote the explicit and the implicit variant respectively.
Because lasso and EN  performed nearly identical we only give the results for the EN.

The results show, that FRI in both variants is superior to \en\ on every data set, especially for clean data where it scores nearly perfect on every measure.
It only shows slightly worse precision in Set 9 where the feature space is big.
\en\ on the other hand, is very precise in that setting, but selects only 37\% of relevant features.
Having shown that, we are now interested in which of the two FRI variants is performing better.
Since they both score perfectly on clean data, we increase the challenge by adding Gaussian noise with a standard deviation of $\sigma = 0.5$ to all sets.
The theory, as given in~\cite{chu_new_2005}, indicates that the implicit variant should perform better on noisy data, because for every decision boundary to be determined it has access to more data samples than the explicit variant, thus gaining an advantage with regard to stability.
However, our experiments do not support this notion as both variants of FRI perform equally well on noisy data.
Interestingly, the \en\ improved its performance on those sets with a lot of weakly relevant features.
This could be explained by assuming that the model has to rely on more of the weak, thus inter-correlated features, to regain the information that was lost due to the introduction of the noise.

\begin{table}[H]
  \caption{Artificially created data sets with known ground truth and evaluation of the identified  relevant features by the methods as compared to all relevant features.
    The data was generated and Gaussian noise (standard deviation $\sigma=0.5$) was added to the predictors. The score is averaged over 30 independent runs. \en\ represents the surrogate model for the ElasticNet with RFECV.}
  \label{t:toyperf_noise}
  \begin{tabular}{l | c || c c c | c c c}
    \hline
    \hline
    \multicolumn{2}{c ||}{}    & \multicolumn{3}{c|}{\textbf{Clean}} & \multicolumn{3}{c}{\textbf{Noise}}                                                             \\
    \hline
    \emph{Metric}              & \emph{Dataset}                      & \en                          & $FRI_{e}$ & $FRI_{i}$ & \en & $FRI_{e}$ & $FRI_{i}$ \\
    \hline
    \multirow{7}{*}{F1}        & Set 1             & 0.94 & 1.0        & 1.0              & 0.92 & 0.95       & 0.98             \\
    & Set 2             & 0.79 & 1.0        & 1.0              & 0.89 & 0.97       & 0.98             \\
    & Set 3             & 0.81 & 1.0        & 1.0              & 0.85 & 0.97       & 0.96             \\
    & Set 4             & 0.83 & 1.0        & 1.0              & 0.80 & 0.96       & 0.97             \\
    & Set 5             & 0.83 & 1.0        & 1.0              & 0.86 & 1.0       & 1.0             \\
    & Set 6             & 0.25 & 1.0        & 1.0              & 0.56 & 0.94       & 0.94             \\
    & Set 7             & 0.49 & 1.0        & 1.0              & 0.46 & 0.90       & 0.91             \\
    & Set 8             & 0.95 & 1.0        & 1.0                 & 0.80 & 0.98       & 0.98                 \\
    & Set 9             & 0.53 & 0.98       & 0.98                 & 0.60 & 1.0       & 1.0                 \\ 
\hline
\multirow{7}{*}{Precision} & Set 1             & 0.90 & 1.0        & 1.0              & 0.87 & 1.0       & 1.0             \\
    & Set 2             & 0.86 & 1.0        & 1.0              & 0.86 & 1.0       & 1.0             \\
    & Set 3             & 0.95 & 1.0        & 1.0              & 0.90 & 1.0       & 1.0             \\
    & Set 4             & 0.95 & 1.0        & 1.0              & 0.91 & 1.0       & 1.0             \\
    & Set 5             & 0.89 & 1.0        & 1.0              & 0.81 & 1.0       & 1.0             \\
    & Set 6             & 1.0 & 1.0        & 1.0              & 1.0 & 1.0       & 1.0             \\
    & Set 7             & 0.97 & 1.0        & 1.0              & 0.84 & 1.0       & 1.0             \\
    & Set 8             & 0.91 & 1.0        & 1.0                 & 0.95 & 1.0       & 1.0                 \\
    & Set 9             & 1.0 & 0.97       & 0.97                & 1.0 & 1.0       & 1.0                 \\ 
\hline
\multirow{7}{*}{Recall}    & Set 1             & 1.0 & 1.0        & 1.0              & 0.99 & 0.92       & 0.96             \\
    & Set 2             & 0.82 & 1.0        & 1.0              & 0.94 & 0.96       & 0.96             \\
    & Set 3             & 0.74 & 1.0        & 1.0              & 0.83 & 0.95       & 0.93             \\
    & Set 4             & 0.77 & 1.0        & 1.0              & 0.74 & 0.93       & 0.94             \\
    & Set 5             & 0.84 & 1.0        & 1.0              & 0.99 & 1.0       & 1.0             \\
    & Set 6             & 0.15 & 1.0        & 1.0              & 0.40 & 0.89       & 0.89             \\
    & Set 7             & 0.41 & 1.0        & 1.0              & 0.35 & 0.84       & 0.86             \\
    & Set 8             & 1.0 & 1.0        & 1.0                 & 0.70 & 0.97        & 0.97                 \\
    & Set 9             & 0.37 & 1.0        & 1.0                 & 0.43 & 1.0       &  1.0                \\
    \hline
    \hline
  \end{tabular}
\end{table}

\subsubsection{Benchmark Data}\label{subs:benchmark_regular}
Here, we purely evaluate the model performance on benchmark data as described in~\cite{zou2005,realOrdinalData} without regarding feature selection.
The imbalanced ordinal regression data sets used in the experiments are listed in Table~\ref{t:three}. 
All samples are normalized to zero mean and unit variance.

\begin{table}[H]
  \caption{Real ordinal regression benchmark data sets with imbalanced classes taken from~\cite{realOrdinalData}, where d is the number of features, and K is the number of classes.}
  \label{t:three}
  \centering
  \begin{tabular*}{\textwidth}{l l || c c c }
    \hline
    \hline
    \textbf{Dataset}           & \textbf{\# Instances} & \textbf{d} & \textbf{K} & \textbf{Ordered Class Distribution}  \\
    \hline
    Automobile        & 205 & 71  & 6 & (3,22,67,54,32,27)      \\
    Bondrate        & 57  & 37  & 5 & (6,33,12,5,1)       \\
    Contact-lenses      & 24  & 6   & 3 & (15,5,4)          \\
    Eucalyptus        & 736 & 91  & 5 & (180,107,130,214,105)   \\
    Newthyroid          & 215 & 5   & 3 & (30,150,35)         \\
    Pasture         & 36    & 25  & 3 & (12,12,12)          \\
    Squash-stored     & 52    & 51  & 3 & (23,21,8)         \\
    Squash-unstored       & 52  & 52  & 3 & (24,24,4)           \\
    TAE           & 151   & 54  & 3 & (49,50,52)          \\
    Winequality-red     & 1599  & 11  & 6 & (10,53,681,638,199,18)  \\

    \hline
    \hline

  \end{tabular*}
\end{table}

We replicate the experiments which have been presented in~\cite{pOGMLVQ,tino} to evaluate the performance of our two possible underlying SVM models as stated in Section~\ref{sec:methods}.
Our models, which we will call \svl[e] and \svl[i] in the following, were tuned using 5-fold cross-validation and used all available features previous feature selection, i.e.\ the models do not use the procedure described in~\ref{sub:handling_numerical_instabilities} and the scores are based on all features without retraining.
The results are averaged over the same 30 folds as used in~\cite{tino} and evaluation is based on the MMAE as defined in Equation~\ref{eq:mmae}.
We compare our models with p-OGMLVQ and a-OGMLVQ, the best performing methods for the given data as stated in~\cite{pOGMLVQ}.
Results for the ElasticNet surrogate \en\ were omitted because they were nearly identical to \svl[e].

The outcomes are reported in Table~\ref{tab:real_ordreg_imp_exp_comparison_mmae_runtime}.
Overall the explicit variant \svl[e] outperforms the implicit variant \svl[i] in all cases except one when considering MMAE.
Similarly, the runtime of \svl[e] is at least two times faster, in some cases even over 20 times faster.
When comparing with the existing results of a-OGMLVQ, we can see \svl[e] outperforming it in 5 cases while being worse in 5 others, it can beat p-OGMLVQ in 6 cases and closely ties in one case (TAE).

\begin{table}[H]
  \caption{Comparison of both proposed variants of ordinal regression models from Section~\ref{sec:methods}.
    Benchmark on real ordinal datasets~\cite{realOrdinalData} by averaged MMAE and aggregated run time over 30 folds. Folds were identical to~\cite{tino} and are comparable.}
  \label{tab:real_ordreg_imp_exp_comparison_mmae_runtime}
  \begin{tabular}{l|cc|cc||cc}
    \hline
    \hline
    {}              & \multicolumn{4}{c||}{\textbf{MMAE}} & \multicolumn{2}{c}{\textbf{Run time}}                                                       \\
    {}              & p-OGMLVQ                            & a-OGMLVQ                              & \svl[e]        & \svl[i]        & \svl[e] & \svl[i] \\
    \hline
    Automobile      & 0.482                               & \textbf{0.446}                        & 0.532          & 0.516          & 151.6   & 876.8   \\
    Bondrate        & 0.768                               & \textbf{0.737}                        & 0.939          & 0.949          & 49.7    & 133.6   \\
    Contact-lenses  & 0.243                               & 0.221                                 & \textbf{0.190} & 0.265          & 23.7    & 53.9    \\
    Eucalyptus      & 0.450                               & 0.477                                 & \textbf{0.390} & \textbf{0.390} & 768.7   & 3280.3  \\
    Newthyroid      & 0.124                               & 0.097                                 & \textbf{0.043} & 0.045          & 37.5    & 92.3    \\
    Pasture         & \textbf{0.307}                      & 0.318                                 & 0.374          & 0.430          & 28.6    & 57.0    \\
    Squash-stored   & 0.415                               & 0.411                                 & \textbf{0.371} & \textbf{0.371} & 36.0    & 68.9    \\
    Squash-unstored & 0.488                               & \textbf{0.228}                        & 0.280          & 0.300          & 35.9    & 69.4    \\
    TAE             & 0.553                               & \textbf{0.537}                        & 0.552          & 0.664          & 43.3    & 83.4    \\
    Winequality-red & 1.078                               & 1.069                                 & 0.868          & \textbf{0.790} & 349.4   & 8359.4  \\
    \hline
    \hline
  \end{tabular}
\end{table}

With regard to feature relevance, no ground truth is available for the given data, rendering us unable to perform the same evaluation as for the artificial sets. 
We are only able to compare the amount of features provided by our method with feature selection (FRI) and the previously used model \en\ as a surrogate for EN with RFECV.
Table~\ref{tab:real_feature_set_size} lists the average number of features identified as relevant for both techniques. 
For three data sets (Squash-stored, Squash-unstored, TAE), FRI identifies a smaller number of relevant features than the alternative, while yielding the same accuracy. 
For three further data sets (Automobile, Eucalyptus, Pasture), FRI identifies more (weakly relevant) features. 
In all cases, FRI potentially offers more information than EN by discriminating between  weakly and strongly relevant features, and giving more candidate features to consider which can than be verified in practise.

\begin{table}[H]
  \caption[Experimental Results]{Mean feature set size of FRI model with explicit constraints and EN surrogate model (\en) with RFECV on real datasets~\cite{zou2005,realOrdinalData}. FRI allows extra discrimination between strong ($FRI^s$) relevance and weak ($FRI^w$) relevance.}
  \label{tab:real_feature_set_size}
  \centering
  \small
  \begin{tabular}{l|ccc|c}
    \hline
    \hline
                    & \multicolumn{4}{c}{\textbf{Average Feature Set Size}}                                    \\ \hline
                    & $FRI_{e}^s$                                           &        & $FRI_{e}^w$ & \en \\ \hline
    Automobile      & 4.5                                                   & $\cup$ & 12.6        & 4.0       \\
    Bondrate        & 0.0                                                   & $\cup$ & 5.4         & 2.0       \\
    Contact-lenses  & 0.9                                                   & $\cup$ & 1.1         & 2.0       \\
    Eucalyptus      & 2.1                                                   & $\cup$ & 33.2        & 15.6      \\
    Newthyroid      & 0.0                                                   & $\cup$ & 4.7         & 2.0       \\
    Pasture         & 0.0                                                   & $\cup$ & 15.5        & 6.0       \\
    Squash-stored   & 2.4                                                   & $\cup$ & 7.9         & 11.1      \\
    Squash-unstored & 1.8                                                   & $\cup$ & 3.3         & 8.0       \\
    TAE             & 1.9                                                   & $\cup$ & 5.4         & 16.8      \\
    Winequality-red & 0.0                                                   & $\cup$ & 7.6         & 5.4       \\
    \hline
    \hline
  \end{tabular}
\end{table}

\subsubsection{COMPAS Analysis}\label{sec:compas}
To showcase a possible application of our approach, we use FRI to examine the COMPAS dataset.
This data was created by Propublica, a journalistic collective from New York, and consists of personal information regarding the criminal history of $11757$ people from Broward County in Florida. 
Data like this has been used to predict an individuals risk of recidivism after a criminal offence. 
Hereby, previous analyses have shown~\cite{compass} that racial bias is incorporated in at least one standard algorithmic prediction tool, meaning that African American individuals receive higher risk scores than Caucasian people.
While it still remains an open research question if and how an algorithm should use socially sensitive attributes~\cite{hajian2013,hardt2016}
we are now interested which information is used by our linear ordinal regression model based on the FRI analysis on the given data.
As such we try to find possible causes for direct or indirect discrimination~\cite{pedreschi} and facilitate careful model design, which seems to be necessary when aiming for long term impact of fair machine learning\cite{liu2018}.

From the originally 28 features of the dataset, we scale down to ten by eliminating all identifying and time related information, which do not contribute information to the prediction task. 
These features are described in detail in~\ref{apx:compas_features}.
We build a predictive model on the data, showing the relevancy of our features to that model.
The result is shown in the upper plot in Figure 1.
In this kind of plot, the relevance intervals are shown as vertical bars such that the maximum and minimum heights represent \textbf{maxrel} and \textbf{minrel}.
For better comparison the values are normalized to the $L_1$ norm of the optimal model ($\|\tilde{\mathbf w}\|_1$).
We also add the maximum element in $\predi(\maxrel)$ as horizontal dashes, which represents the threshold which is used to classify between weakly relevant and irrelevant features. 

The predictive accuracy is $66.73 \%$ which is directly inside the range of accuracies discussed in the Propublica analysis - note that the models used in practice deviate from the ones considered here, and the former are not available to us. 
Thus, we discuss properties of the linear models found by the proposed ORP only, not any other model.
Two features are strongly relevant, namely, the count of prior charges and the age group 17-25 which show a big contribution in absolute terms.
Many other features, such as the count of juvenile felonies and misdemeanors, or the degree of criminal changes are weakly relevant.
More interestingly, socially sensitive features such as the sex and race are also considered weakly relevant.
In the case of sex, both male and female exhibit the same maximal relevance which hints at the anti-correlation between the two features.
In the case of race, being African-American, Caucasian or Native American is considered weakly relevant.
When compared with the Propublica analysis, our relevance bounds are in line with their results.

To measure the contribution of the ethnic features in the model, we repeat the experiment with all those features removed.
Hereby, the accuracy does not drop significantly, yielding $65.99 \%$. 
The bottom plot of Figure 1 shows the relevance for all remaining features. 
Compared to the previous model, there are two notable changes.
The count of juvenile offences and the information about violent recidivism become relevant which are intuitively much more important to the problem at hand and do not reiterate a potential bias in society.

\begin{figure}[H]
  \label{fig:compas1}
  \minipage{\textwidth}
  \includegraphics[width=\linewidth]{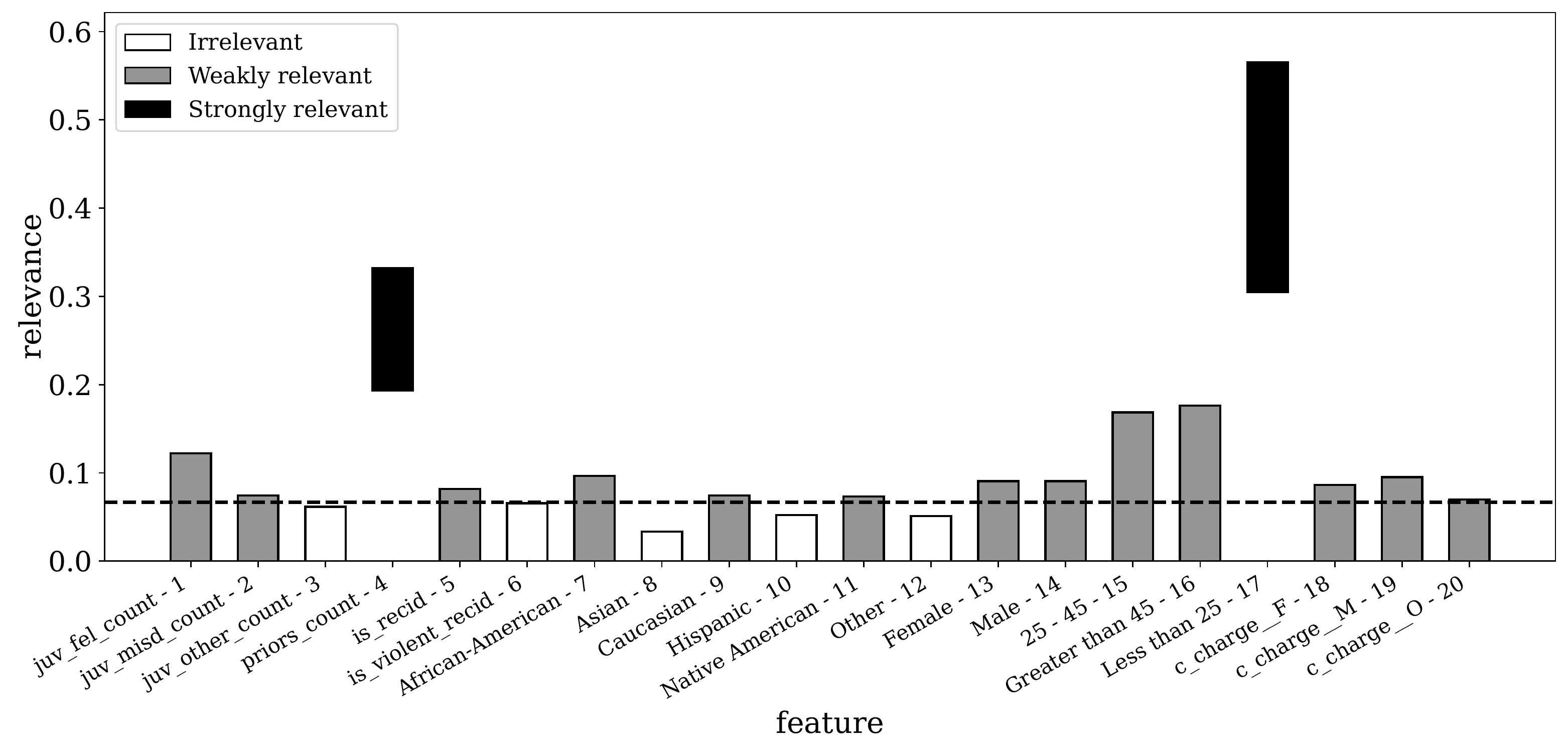}
  \label{fig:all_variables}
  \endminipage \\
  \minipage{\textwidth}
  \includegraphics[width=\linewidth]{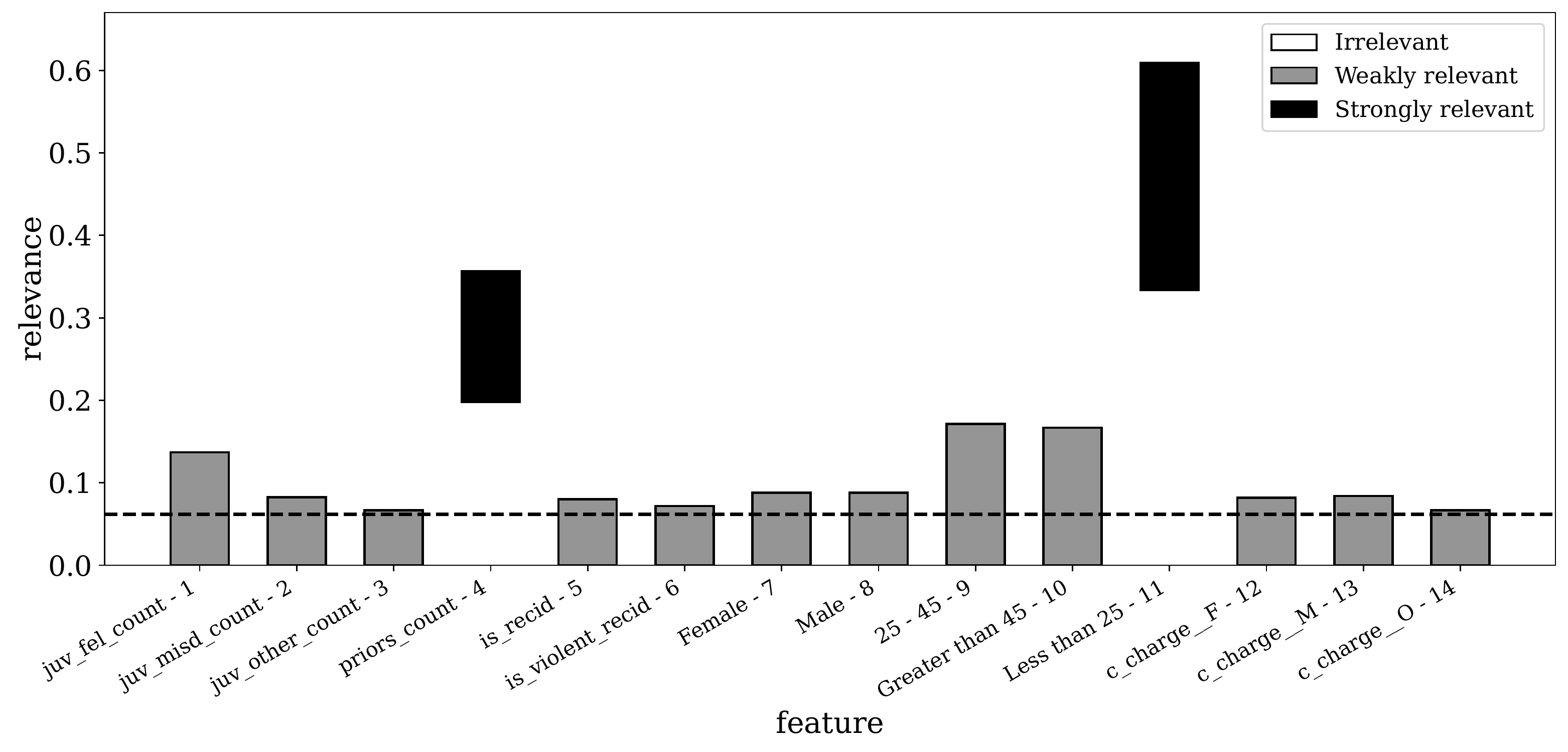}
  \label{fig:f_m_no_race}
  \endminipage
  \caption[Plots of the COMPAS dataset]{Relevance plots for the COMPASS dataset. Top: Relevance intervals (bars) for all features including ethnicity. Bottom: Relevance intervals for all features when ethnicity is eliminated from the data. Ethnicity is not a relevant factor for the model on top, so if those variables are eliminated, the relevancy of the other features do not change profoundly. The y-axis represents the computed feature relevance normalized to the $L_1$ norm of the optimal model.}
\end{figure}

\subsection{Privileged Information}\label{sub:experiment_privileged_information}
The following section evaluates our approach for the LUPI paradigm, i.e.\  our method handling privileged information, that we denote $FRI^*$. From here, we focus on the explicit variant, after showing its superiority over the implicit version in section~\ref{subs:benchmark_regular} as regards computational complexity, leading to the notation $FRI_e^*$.  Again, we show the quality of our feature selection by testing on artificially created data with known ground truth. Due to a lack of specific LUPI benchmark datasets, we conclude our paper with a semantic analysis of a $FRI_e^*$ model on one demonstrative example.

\subsubsection{Artificial Data}

We use the generation method presented in~\cite{lopez-paz2015} to create artificial datasets containing regular as well as privileged information by sampling triplets $(x_i , x_{i}^{*}, y_i)$ from:

\begin{equation*}
  x_i^* \sim \mathcal{N}(0, I_{d})
\end{equation*}
\begin{equation*}
  \epsilon_i \sim \mathcal{N}(0, I_{d})
\end{equation*}
\begin{equation*}
  x_i \gets x_i^* + \epsilon
\end{equation*}
\begin{equation*}
  y_i \gets f(\langle \omega , x_i^* \rangle),
\end{equation*}

where $f$ denotes a function that assigns the correct ordinal bin to the label $y_i$ based on the value of the dot product between the weight vector and a privileged sample $x_i^*$.

Hereby, the privileged information $X^*$ consists of clean versions of the noisy regular features $X$. 
Both, the regular and the privileged feature space, contain strong, weak and irrelevant features. 
These are created in the same way as described in section~\ref{subs:artif_data}. 
The characteristics of the data used in our experiments are shown in Table~\ref{t:artificial_lupi}. 
The last two sets differ from the generation method mentioned above.
Their regular information is created similarly to the sets in Table~\ref{t:artificial_data}, to which three irrelevant privileged features are added from random Gaussian noise. 
All features are normalized to zero mean and unit variance.

\begin{table}[H]
  \caption{Artificially created data with regular and privileged features under known ground truth.
    For the first six sets, the privileged features consist of clean versions of the regular information.
    The last two sets are regular ordinal regression sets with random noise as additional privileged information.}
  \label{t:artificial_lupi}
  \centering
  \begin{tabular*}{\textwidth}{c c || ccc | ccc}
    \hline
    \hline
    &   & \multicolumn{3}{c|}{\textbf{Regular Features}} & \multicolumn{3}{c}{\textbf{Privileged Features}}  \\
    \hline
    \emph{Dataset} & \emph{\#Instances}   & \emph{\#Str} & \emph{\#Weak} & \emph{\#Irr} & \emph{\#Str} & \emph{\#Weak} & \emph{\#Irr} \\
    \hline
    Set 1 					& 200 & 6 & 0 & 3 & 6 & 0 & 3    	 \\
    Set 2					& 200 & 0 & 12 & 3 & 0 & 12 & 3 	 \\
    Set 3				    & 200 & 6 & 6 & 0 & 6 & 6 & 0		 \\
    Set 4					& 200 & 3 & 6 & 0 & 3 & 6 & 0		 \\
    Set 5		    	 	& 200 & 1 & 4 & 0 & 1 & 4 & 0		 \\
    Set 6					& 200 & 1 & 40 & 10 & 1 & 40 & 10		 \\
    \hline
    Set 7					& 200 & 4 & 2 & 2 & 0 & 0 & 3      \\
    Set 8                   & 200 & 0 & 4 & 2 & 0 & 0 & 3      \\

    \hline
    \hline

  \end{tabular*}
\end{table}

Evaluation closely follows section~\ref{subs:artif_data}. 
Again, we use the F-measure as a quantifying metric for the detection of the all relevant features set, and compare our method to the EN surrogate model \en. 
While $FRI_e^*$ differentiates between the two feature spaces in the data, the EN receives both the regular and the privileged set as one. 
With that, we want to showcase the advantages of a LUPI model for feature selection over a purely regular model.

The results are given in Table~\ref{lupi_perf}. 
$FRI_e^*$ achieves a perfect score on the regular feature set and only stumbles once, for set 6, on the privileged information. 
The EN on the other hand, performs considerably worse on the regular set but shows significant improvements on the privileged set, albeit it cannot match the performance of our method. 
The improvements on the privileged data are easy to explain since this information is the clear original information as opposed to the noisy features in the regular set.

\begin{table}[H]
  \caption{Artificially created datasets with known ground truth and evaluation of the identified relevant features by the methods as compared to all existing relevant features. The EN surrogate model (\en) receives both feature sets as one but evaluation is done separately for the regular and privileged feature set. The score is averaged over 10 independent runs.}
  \label{lupi_perf}
  \begin{tabular}{l c  || cc | cc}
    \hline
    \hline
                               &                & \multicolumn{2}{c|}{\textbf{Regular Features}} & \multicolumn{2}{c}{\textbf{Privileged Features}}                           \\
    \hline
    \emph{Metric}              & \emph{Dataset} & \en                                      & $FRI_{e}^*$                                      & \en & $FRI_{e}^*$ \\
    \hline
    \multirow{8}{*}{F1}        & Set 1          & 0.44                                           & 1.0                                              & 0.89      & 1.0         \\
                               & Set 2          & 0.48                                           & 1.0                                              & 0.85      & 1.0         \\
                               & Set 3          & 0.65                                           & 1.0                                              & 0.91      & 1.0         \\
                               & Set 4          & 0.58                                           & 1.0                                              & 0.88      & 1.0         \\
                               & Set 5          & 0.67                                           & 1.0                                              & 0.92      & 1.0         \\
                               & Set 6          & 0.40                                           & 1.0                                              & 0.69      & 0.99        \\
                               & Set 7          & 0.93                                           & 1.0                                              & 1.0       & 1.0         \\
                               & Set 8          & 0.70                                           & 1.0                                              & 1.0       & 1.0         \\
    \hline
    \multirow{8}{*}{Precision} & Set 1          & 0.72                                           & 1.0                                              & 0.91      & 1.0         \\
                               & Set 2          & 0.75                                           & 1.0                                              & 0.98      & 1.0         \\
                               & Set 3          & 1.0                                            & 1.0                                              & 1.0       & 1.0         \\
                               & Set 4          & 0.90                                           & 1.0                                              & 1.0       & 1.0         \\
                               & Set 5          & 0.80                                           & 1.0                                              & 1.0       & 1.0         \\
                               & Set 6          & 0.98                                           & 1.0                                              & 0.97      & 1.0         \\
                               & Set 7          & 0.94                                           & 1.0                                              & 1.0       & 1.0         \\
                               & Set 8          & 1.0                                            & 1.0                                              & 1.0       & 1.0         \\
    \hline
    \multirow{8}{*}{Recall}    & Set 1          & 0.37                                           & 1.0                                              & 0.88      & 1.0         \\
                               & Set 2          & 0.38                                           & 1.0                                              & 0.78      & 1.0         \\
                               & Set 3          & 0.52                                           & 1.0                                              & 0.84      & 1.0         \\
                               & Set 4          & 0.48                                           & 1.0                                              & 0.80      & 1.0         \\
                               & Set 5          & 0.62                                           & 1.0                                              & 0.88      & 1.0         \\
                               & Set 6          & 0.26                                           & 0.99                                             & 0.54      & 0.98        \\
                               & Set 7          & 0.93                                           & 1.0                                              & 1.0       & 1.0         \\
                               & Set 8          & 0.55                                           & 1.0                                              & 1.0       & 1.0         \\
    \hline
    \hline
  \end{tabular}
\end{table}

\subsubsection{Semantic Analysis}
Performing evaluations similar to sections~\ref{subs:benchmark_regular} and~\ref{sec:compas} is not possible because of the lack of public LUPI benchmark.
Therefore, we consider one illustrative example to demonstrate the semantic implications of the FRI framework for LUPI.
We generate a set with $400$ samples and six features. 
Initially, there are three strongly relevant features and three irrelevant ones drawn from random Gaussian noise. 
We divide the samples into four groups, each with $100$ members. 
The first group has Gaussian noise with a standard deviation of $0.1$ added to the first strongly relevant feature. 
The second group has a noise level of $0.5$ added to the second feature. 
Similarly, the third one, has Gaussian noise on the last strong feature with a standard deviation of $2$. 
The data in the last group is noise free. 
The idea is to provide the insight which samples of the dataset are hard to classify as privileged information to the model. 
Therefore, the privileged set consists of three features, incorporating the noise that was added to the groups, with the first privileged feature corresponding to the first group and so on. 

The plots in Figure 2 show the relevancy for the regular features (a) as well as for the privileged features (b). 
Our method correctly dismissed the three irrelevant features and also classifies all strongly relevant features.
More importantly, all privileged features were also correctly classified, and their relevance correlates with the noise level.
With that, we show that $FRI_e^*$ can discriminate between the usefulness of multiple privileged features and utilize those that are necessary in this setting.

\begin{figure}[H]
  \label{fig:lupi}
  \includegraphics[width=\linewidth]{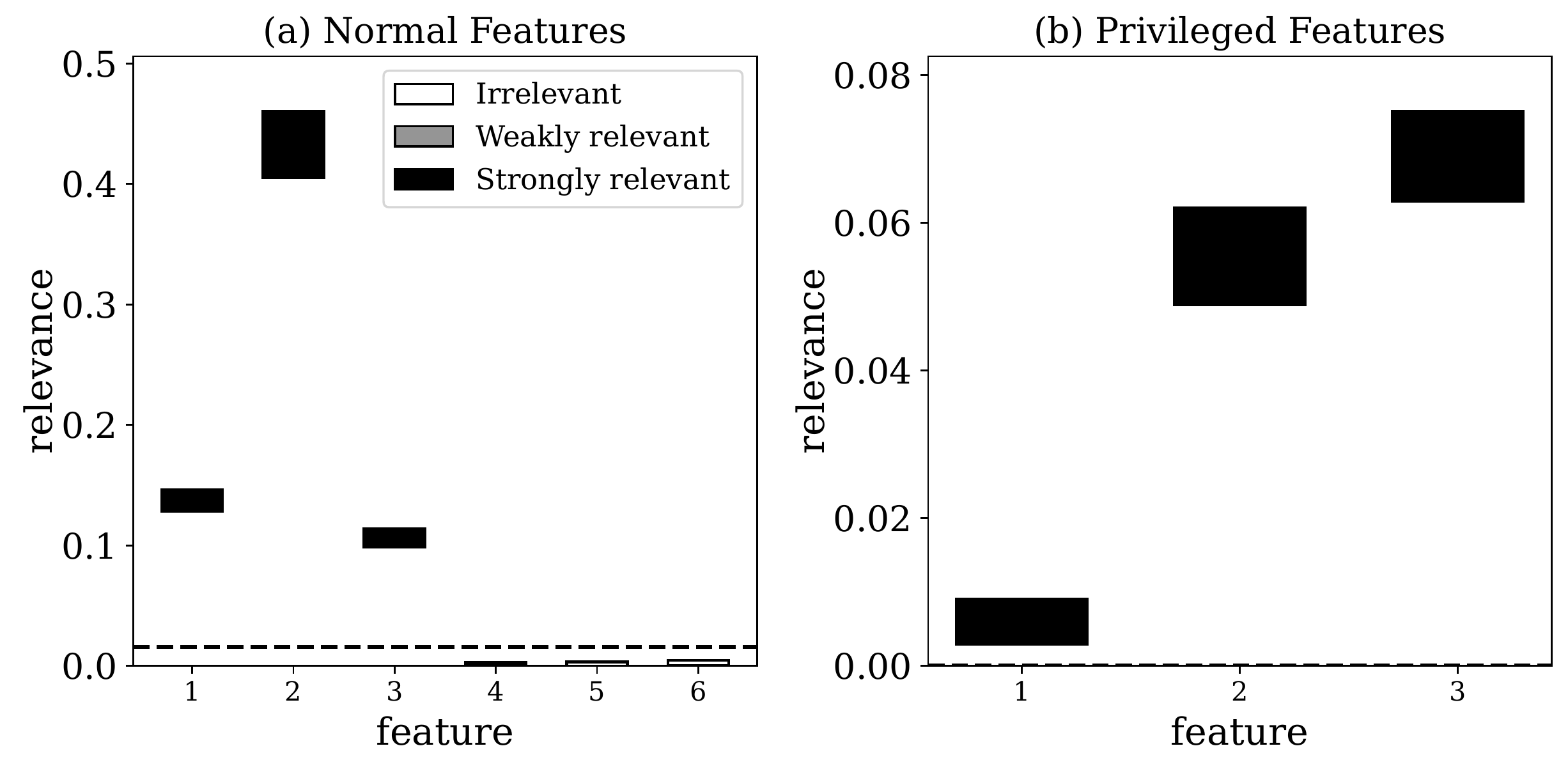}
  \caption[Plots of the LUPI model.]{Relevance plots for the semantic analysis. 
   (a) Relevance of the regular features for the LUPI model.
   (b): Relevance of the privileged features for the LUPI model.}
\end{figure}

\section{Conclusions}\label{sec:conclusion}

In this paper we presented the adaption of the feature relevance bounds approach to ordinal regression data using the \emph{explicit} order variant. The optimization problem was phrased by approximating the generalization ability of the model with a bound on the $L_1$-margin. The resulting problem can be transferred to a linear problem. For its solution, we used another approximation by splitting the objective into the margin and slack variables separately, for larger robustness.
Further, we proposed a resampling-based procedure to determine which values correspond to no information of the features, to automatically set situation-dependent thresholds.
Based on the experiments we showed that the \emph{explicit} variant is comparable to the \emph{implicit} variant for this use case on the given data as regards the accuracy and more efficient.
Our method can provide a near perfect all-relevant feature set approximation while being significantly faster than the other variant.
Although not many feature selection approaches exist for that specific context we could also showcase the feature selection performance in comparison with another popular approach on toy and real data.
The feature sets produced by our approach represents additional information useful in analytic use cases for model and experiment design, subject for further evaluation, and it constitutes a possible starting point to investigate, e.g.\ the information which restricted or protected features can provide for the class of linear ORP models.

Furthermore, we also provided a definition for feature relevance bounds when additional information is present in the context of learning using privileged information.
Here we defined a features relevant in relation to the training phase itself.
Similar to the classical context, our method achieved very good feature selection sensitivity in both the regular and privileged feature set, this way enabling a strategy to choose suitable features or teacher information to facilitate training.


\footnotesize

\bibliographystyle{elsarticle-num}
\bibliography{neurocomputing,revision}

\appendix
\section{Proof of Theorem~\ref{thm1}}
We proof Theorem~\ref{thm1}. We rely on
Theorem 4 in~\cite{christina}, which states the following:
Assume two optimization problems
\begin{eqnarray}\notag
  &&\mathrm{Problem\ A: }\min_x h_1(x) \mathrm{\ s.t.\ }  x\in A_1\\\notag
  &&\mathrm{Problem\ B: }\min_y h_2(y) \mathrm{\ s.t.\ }  y\in A_2\\\notag
\end{eqnarray}
Assume mappings $f:A_1\to A_2$ and $g:A_2\to A_1$ exist such that for all $x\in A_1$, $y\in A_2$
\begin{eqnarray*}
  h_2(y)<h_2(f(x)) & \Rightarrow & h_1(g(y))<h_1(x)\\
  h_1(x)<h_1(g(y)) & \Rightarrow & h_2(f(x))<h_2(y)
\end{eqnarray*}
Then the two problems $A$ and $B$ are equivalent in the sense that the mappings $f$ and $g$ establish
direct correspondences of their global optima.

\subsubsection*{Equivalence of $\mathrm{minrel}(\feature)$ and $\mathrm{minrel}^*(\feature)$}

Solutions of $\mathrm{minrel}(\feature)$ have the form \[\mathbf{w}=(w_1,\ldots,w_n),b=(b_1,\ldots,b_{l-1}),\chi=(\chi_1^1,\ldots,\chi_{m_{l-1}}^{l-1}),\xi=(\xi_1^2,\ldots,\xi_{m_l}^l))\]
$\mathrm{minrel}^*(\feature)$ combines this form with an additional vector
$\hat{\mathbf{w}}=(\hat w_1,\ldots,\hat w_n)$.
Define the mapping \[f:(\mathbf{w},b,\chi,\xi)\mapsto(\mathbf{w},\hat{\mathbf{w}}=|\mathbf{w}|:=(|w_1|,\ldots,|w_n|),b,\chi,\xi)\]
Define the mapping \[g: (\mathbf{w},\hat{\mathbf{w}},b,\chi,\xi)
  \mapsto(\mathbf{w},b,\chi,\xi) \]
$f$ is obviously a mapping in between feasible sets. The same holds for $g$, since the constraints (\ref{eq9}) ensure
$\sum_k\hat w_k\ge \|\mathbf{w}\|_1$.

Given an element of the feasible set of the two problems, denoted by
$x:=(\mathbf{w}^A,b^A,\chi^A,\xi^A)$ and $y:=(\mathbf{w}^B,\hat{\mathbf{w}}^B, b^B,\chi^B,\xi^B)$, respectively.
Assume $h_2(y)<h_2(f(x))$, i.e.\ $\hat{w}_\feature^B<|w_{\feature}^A|$. Then constraints (\ref{eq9}) ensure $|w_{\feature}^B|\le \hat{w}_\feature^B$, hence
$|w_{\feature}^B| < |w_{\feature}^A|$, i.e.\ $h_1(g(y))<h_1(x)$.

Conversely, $h_1(x)<h_1(g(y))$ implies $|w_{\feature}^A|<|w_{\feature}^B|$ hence constraints (\ref{eq9}) ensure
$|w_{\feature}^A| < \hat w_{\feature}^B$, i.e.\ $h_2(f(x))<h_2(y)$.

\subsubsection*{Equivalence of $\mathrm{maxrel}(\feature)$ and the optimum of $\mathrm{maxrel}^*_{\mathrm{pos}}(\feature)$
and $\mathrm{maxrel}^*_{\mathrm{neg}}(\feature)$}

We consider two problems which are associated to $\mathrm{maxrel}(\feature)$,
$\mathbf{maxrel}_{\mathrm{pos}}(\feature)$ equals $\mathrm{maxrel}(\feature)$ with the additional constraint $w_{\feature}\ge 0$, and
$\mathbf{maxrel}_{\mathrm{neg}}(\feature)$ equals $\mathrm{maxrel}(\feature)$ with the additional constraint $w_{\feature}\le 0$.
Since these two auxiliary problems decompose the feasible set of the original one into two halves, we can
solve these two auxiliary problem and take whichever solution is best
instead of solving  $\mathrm{maxrel}(\feature)$. Thus, we can show equivalence of these two sub problems to the versions as introduced in Theorem (\ref{thm1}). Instead of maximization, we can focus on the minimization of the respective negative of the original objectives, to phrase the setting within the notation of Theorem 4 in~\cite{christina}.

We show equivalence of
$\mathrm{maxrel}_{\mathrm{pos}}(\feature)$ and  $\mathrm{maxrel}_{\mathrm{pos}}^*(\feature)$.
Define the mapping $f$ as identity for $(\mathbf{w},b,\chi,\xi)$ and $\hat{\mathbf{w}}=|\mathbf{w}|:=(|w_1|,\ldots,|w_n|)$.
Define the mapping $g$ as projection of $(\mathbf{w},\hat{\mathbf{w}},b,\chi,\xi)$ onto all elements but $\hat{\mathbf{w}}$.
$f$ and $g$ are obviously mappings in between the feasible sets. Note that constraints $w_{\feature}\ge 0$ and $\hat w_{\feature}\le w_{\feature}$ are
required at this step.

Given elements of the feasible sets of the problems $x:=(\mathbf{w}^A,b^A,\chi^A,\xi^A)$ and $y:=(\mathbf{w}^B,\hat{\mathbf{w}}^B, b^B,\chi^B,\xi^B)$.
Assume $h_2(y)<h_2(f(x))$, i.e.\ $-\hat{w}_\feature^B<-|w_{\feature}^A|$. Then the constraints (\ref{eq9}) and (\ref{eq11})  ensure
$\hat w_{\feature}^B = w_{\feature}^B$ and $\hat w_{\feature}^B\ge 0$, hence
$-|w_{\feature}^B| < -|w_{\feature}^A|$, i.e.\ $h_1(g(y))<h_1(x)$.

Conversely, $h_1(x)<h_1(g(y))$ implies $-|w_{\feature}^A|<-|w_{\feature}^B|$ hence
$-|w_{\feature}^A| < -\hat w_{\feature}^B$, i.e.\ $h_2(f(x))<h_2(y)$ due to constraints (\ref{eq9}) and (\ref{eq11}).

Similarly, equivalence of
$\mathrm{maxrel}_{\mathrm{neg}}(\feature)$ and  $\mathrm{maxrel}_{\mathrm{neg}}^*(\feature)$ can be shown.
$f$ and $g$ are as above. These are mappings in between feasible sets.

$f$ and $g$ are obviously mappings in between the feasible sets. Note that constraints $w_{\feature}\ge 0$ and $\hat w_{\feature}\le w_{\feature}$ are
required at this step.

\section{Feature Relevance Bounds for Ordinal Regression with Implicit Order}\label{apx:rele_bounds_implicit}
In the following we are defining the relevance bounds for the implicit variant from Section~\ref{sec:methods}.
The definition is very similar to Section~\ref{sec:relev_bounds}, and the following will be very concise.

Assume a training set $X$.
Denote an optimum solution of problem (\ref{eq:svm_ordreg_imp}) as  $(\tilde{\mathbf w},\tilde b_j,\tilde \xi_i^j, \tilde \chi_i^j)$.
We define \[
  L:= \sum_{j=1}^{l-1}
  \left( \sum_{k=1}^{j}\sum_{i=1}^{n^k} \chi_{ki}^j +
  \sum_{k=j+1}^{l}\sum_{i=1}^{n^k}\xi_{ki}^j
  \right)
\] as the sum of all slack variables.
The optimum solution induces the
value
\[
  \mu_X:=\frac{1}{2}\|\tilde{\mathbf w}\|_1 + C \cdot L
\] which is uniquely determined by $X$.

The class of equivalent good hypotheses is proxied by

\begin{eqnarray}\notag
  F_{\delta}(X)&:=&\{\mathbf w\in\mathbb{R}^n\:|\:\exists \bm \xi, \bm \chi, \mathbf b \mbox{ such that constraints in (\ref{eq:svm_ordreg_imp}) hold,}\\\notag
  && \frac{1}{2}\|\mathbf w\|_1+ C\cdot L \le (1+\delta)\cdot \mu_X\}
\end{eqnarray}

\begin{description}
  \item[Problem $\mathbf{minrel}(\feature)$:]
        \begin{eqnarray}
          \min_{\mathbf w,\mathbf b,\bm \chi, \bm \xi} && |w_{\feature}|\\\notag
          \mbox{s.t.\ for all } i,j&& \mbox{conditions in (\ref{eq:svm_ordreg_imp}) hold}\\
          &&\frac{1}{2}\|\mathbf{w}\|_1 + C\cdot L \le (1+\delta)\cdot\mu_X
        \end{eqnarray}

  \item[Problem $\mathbf{maxrel}(\feature)$:]
        \begin{eqnarray}
          \max_{\mathbf w,\mathbf b,\bm \chi, \bm \xi} && |w_{\feature}|\\\notag
          \mbox{s.t.\ for all } i,j&& \mbox{conditions in (\ref{eq:svm_ordreg_imp}) hold}\\
          &&\frac{1}{2}\|\mathbf{w}\|_1 + C\cdot L \le (1+\delta)\cdot\mu_X
        \end{eqnarray}
\end{description}

As before, this problem can be equivalently phrased as a linear program.

\subsection{Scaling}\label{sec:scaling_eval}
    Here we evaluate the scaling of our implementation in the setting without privileged information.
    We already discussed the theoretical time complexity bounds in Section~\ref{sub:complexity} where we concluded that the overall method with feature selection is in \bigO{n^3 + (dz+c)n^{2.5}}.
    We now run two separate experiments where we generate artificial sets as described earlier and scale up their size with respect to the number of instances $n$ and number of features $d$.
    In the first experiment we set $d=20$ and scale $n$ between $10$ and $10000$ and in the second we set $n=500$ and scale $d$ between $10$ and $500$.
    Our implementation is using the high level library \emph{cvxpy}\footnote{\href{https://www.cvxpy.org}{https://www.cvxpy.org}} and the ECOS solver~\cite{Domahidi2013Ecos} and presents no specific adaption for the problems at hand.
    The implementation runtime is measured on a modern Intel Xeon processor.
    Additionally, because relevance bounds can be computed in parallel, we run both experiments with one single thread and 8 threads in parallel.

\begin{figure}
    \centering
    \begin{subfigure}[b]{0.4\textwidth}
        \includegraphics[width=\textwidth]{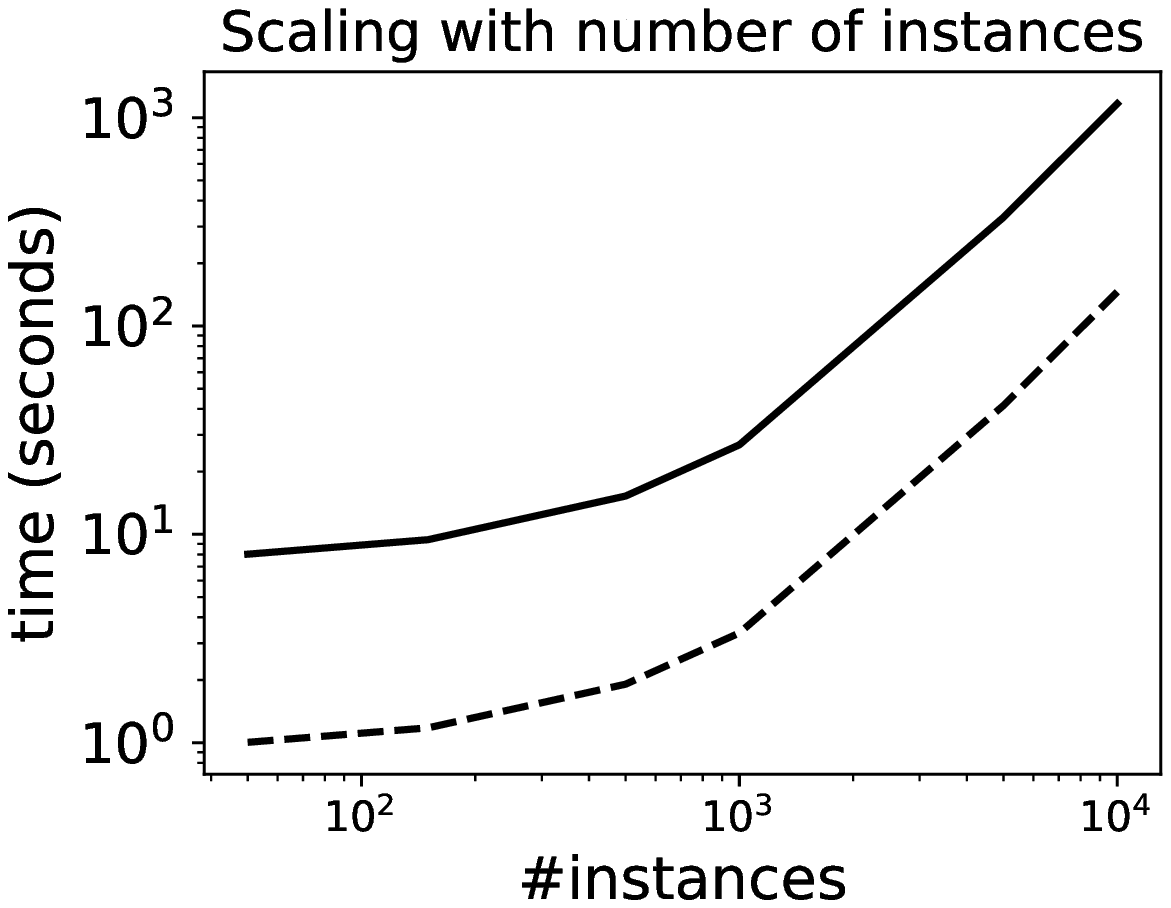}
        \caption{Instances}
        \label{fig:time_instances}
    \end{subfigure}
    \quad 
    \begin{subfigure}[b]{0.4\textwidth}
        \includegraphics[width=\textwidth]{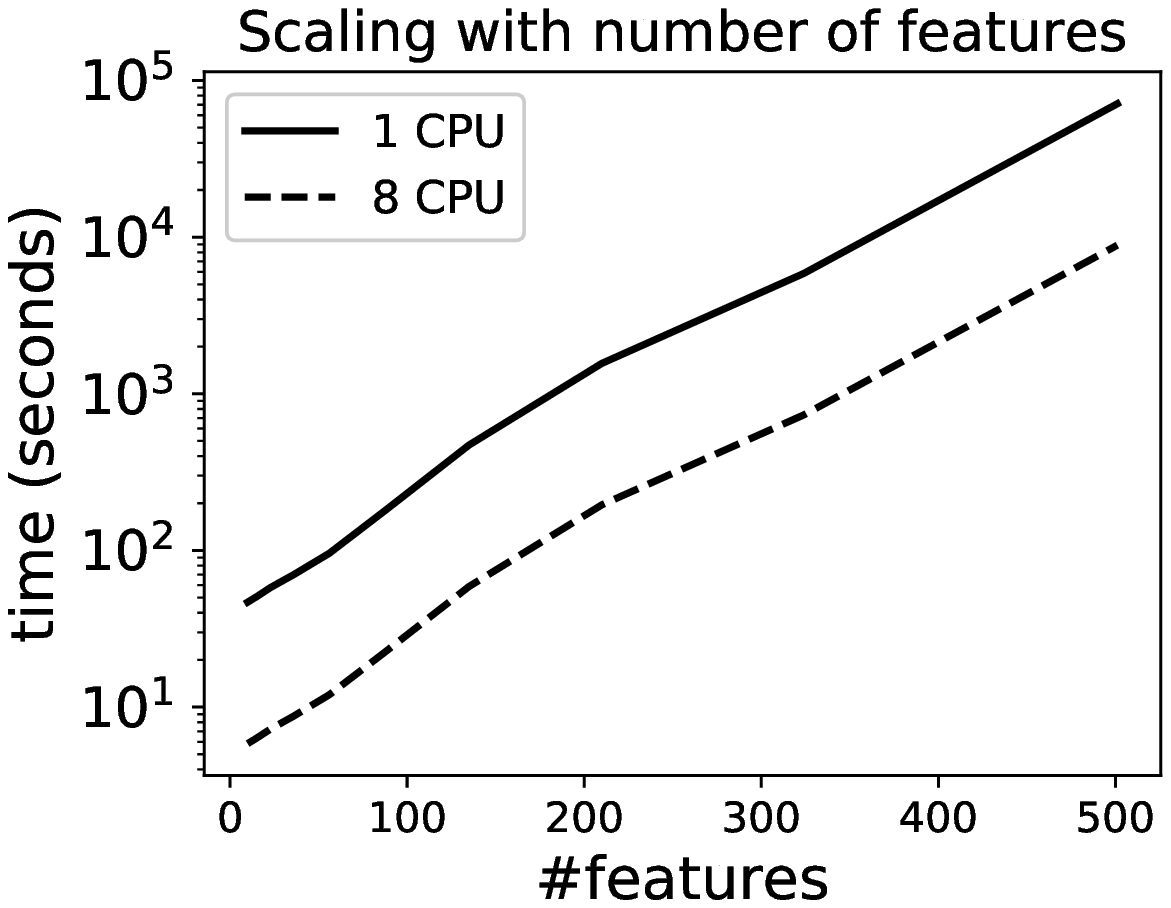}
        \caption{Features}
        \label{fig:time_features}
    \end{subfigure}
    \caption{Plot of runtime with respect to number of instances (a) and number of features (b). Additionally, both show comparison between single thread (1 CPU) and multi threaded run (8 CPU).
          }\label{fig:runtimes}
\end{figure}

    In Figure~\ref{fig:runtimes} results for both experiments are given.
    One can see that the complexity can limit the application of the method to small to medium-sized problems.
    This is in line with other \emph{all} relevant feature selection methods~\cite{pfannschmidt2019a} which exhibit much higher runtimes than simple sparse methods.
    While slightly bigger sets with, e.g. $n>10^4$ or $d>500$ are feasible, multiprocessing is recommended.
    For bigger data sets, further optimization or filtering of the feature space is necessary.

\section{Features of the COMPAS dataset}\label{apx:compas_features}

This section gives a description of all the features of the COMPAS dataset that we use in our analysis in~\ref{sec:compas}. The features are listed in Table~\ref{t:compas}. All categorical variables are One-Hot-Coded for the analysis. The ethnicities are one of \{African-American, Caucasian, Hispanic, Asian, Native American, Other\}, the sexes are male or female, the age is grouped into \{less than 25, 25-45, greater than 45\} and the charge can be one of \{felony, misdemeanour, offence\}. The total number of features fed into the first model is 20. After eliminating all ethnic information, the count reduces to 14.

\begin{table}[H]
  \caption{Description of features of the COMPAS dataset used for the analysis in~\ref{sec:compas}.}\label{t:compas}
  \centering
  \begin{tabular*}{\textwidth}{l | c | l | c } \\
    \hline
    \hline
    \textbf{Feature Name} & \textbf{Type} & \textbf{Description} & \textbf{One-Hot}  \\
    \hline
    Juv\_fel\_count      &   Numerical     &   \# Felonies as a juvenile   &   No     \\
    Juv\_misd\_count      &   Numerical     &   \# Misdemeanour as a juvenile   &   No     \\
    Juv\_other\_count &   Numerical   &   \# Offences as a juvenile   &   No     \\
    Priors\_count           &   Numerical    &   \# Prior convictions    &   No    \\
    Is\_recid    &   Binary  &   If recidivism happened   &   No      \\
    Is\_violent\_recid    &   Binary  & If violent recidivism happened &      No      \\
    Ethnicity   &   Categorical     & One of 6 ethnicities  &   Yes \\
    Sex         &   Categorical     & One of 2 sexes        &   Yes \\
    Age         &   Categorical     & One of 3 age groups   &   Yes \\
    C\_charge    &   Categorical     & One of 3 charge groups    & Yes \\

    \hline
    \hline

  \end{tabular*}
\end{table}

\end{document}

%% file: mymath.tex
\newcommand{\IR}{\mathbb{R}}

\renewcommand{\epsilon}{\varepsilon}

\newcommand{\norm}[1]{\left\lVert #1 \right\rVert}

\newcommand{\onenorm}[1]{\norm{#1}_1}

\newcommand{\regparameter}{\mathit{C}}
\newcommand{\slack}{\xi}

\newcommand{\costsum}[1]{\ensuremath{\sum_{i=1}^{n} \slack{}_i}}

\newcommand{\svl}[1][]{$\text{M}^{L_1}_{#1}$}
\newcommand{\en}{$\text{M}^{L_{1}+L_{2}}_{e}$}

\newcommand{\feature}{\mathcal{I}}
\newcommand{\pfeature}{\mathcal{P}}

\newcommand{\xp}[1][]{\mathbf x^{*{#1}}_i}   
\newcommand{\Xp}{ X^*}
\newcommand{\dpriv}[1][j]{d_{#1}}
\newcommand{\omegap}[1][]{\mathbf{w}^*_{#1}}

\newcommand{\wph}[1][\feature]{\hat w^{*}_{#1}}
\newcommand{\lonesum}[1][w]{\sum_k #1}
\newcommand{\loneconstraint}[2]{{#1}_{#2} \le \hat {#1}_{#2}, \, -{#1}_{#2} \le  \hat {#1}_{#2} }

\newcommand{\fprivchi}[1][j]{p^{#1}_{\chi}(\xp[])}
\newcommand{\fprivxi}[1][j]{p^{#1}_{\xi}(\xp[])}
\newcommand{\privfuncxi}{\left( \mathbf{w}^{*}_{\xi} \cdot \xp[j]+ d_{\xi} \right)}
\newcommand{\privfuncchi}{\left( \mathbf{w}^{*}_{\chi} \cdot \xp[j]+ d_{\chi} \right)}
\newcommand{\wnorm}[1][w]{\frac{1}{2} \onenorm{\mathbf {#1} }}
\newcommand{\wprivnorm}{\frac{\gamma}{2} ( \onenorm{\mathbf{w}^{*}_{\chi}}+\onenorm{\mathbf{w}^{*}_{\xi}} ) }
\newcommand{\privloss}{ C \sum_{j=1}^{l} \sum_{i=1}^{n^k} \left(\fprivchi + \fprivxi \right)}
\newcommand{\privlossfunc}{\mathcal{L}}
\newcommand{\optwprivnorm}{\frac{\gamma}{2} ( \onenorm{\tilde{\mathbf{w}}_{\chi}}+\onenorm{\tilde{\mathbf{w}}_{\xi}} )}
\newcommand{\svmpriv}{\wnorm + \wprivnorm + \privloss }

\newcommand{\optsvmpriv}{\wnorm[\tilde{w}] + \optwprivnorm + \tilde{\privlossfunc}} 
\newcommand{\privproxy}{F_{\delta}(X,\Xp)}
\newcommand{\privmu}{\mu_{X,\Xp}}

\newcommand{\maxrel}{\text{maxrel}}
\newcommand{\minrel}{\text{minrel}}
\newcommand{\perm}[1]{p(#1)}

\newcommand{\datawithperm}{X^{\perm \feature}}
\newcommand{\probedist}[1]{\widehat{\pi}(#1)}
\newcommand{\predi}{\Pi}
\newcommand{\PI}[1]{\predi(#1)_n := \displaystyle {\overline {\probedist{\cdot}}}_{n}\pm T_{n-1}(p) \cdot \sigma(\probedist{\cdot}) \sqrt {1+(1/n)}}

\newcommand{\bigO}[1]{$\mathcal{O}(#1)$}